\documentclass[a4paper,fleqn]{cas-dc}
\usepackage{natbib}
\usepackage{amsmath,amsfonts}
\usepackage{graphicx}
\usepackage{booktabs}
\usepackage{longtable}
\usepackage{float}
\usepackage{makecell}
\usepackage{multirow}
\usepackage{xcolor,colortbl}
\usepackage{array}
\usepackage{subcaption}  % 子图
\usepackage{algorithm}
\usepackage{algpseudocode}
\usepackage{hyperref}    % 只加载一次，若报 option clash 就把这行删掉
\definecolor{darkergreen}{RGB}{0,100,0}
\usepackage{adjustbox}
\setcitestyle{numbers,square,sort&compress}
\begin{document}

\let\WriteBookmarks\relax
\def\floatpagepagefraction{1}
\def\textpagefraction{.001}

\shorttitle{} 
\shortauthors{}

\title[mode=title]{Generating Transferrable Adversarial Examples via Local Mixing and Logits Optimization for Remote Sensing Object Recognition}

% 作者 + 通讯作者
% 作者 + 通讯作者
\author[inst1]{Chun Liu}
\ead{liuchun@henu.edu.cn}

\author[inst2]{Hailong Wang}
\author[inst2]{Bingqian Zhu}
\author[inst2]{Panpan Ding}

\author[inst3]{Zheng Zheng}
\ead{zhengz@buaa.edu.cn}

\author[inst2]{Tao Xu}\cormark[1]   % 在名字后面加 \corref
\ead{txu@henu.edu.cn}

\author[inst1]{Zhigang Han}
\ead{zghan@henu.edu.cn}

\author[inst1]{Jiayao Wang}
\ead{wjy@henu.edu.cn}

% 通讯作者说明
\cortext[cor1]{Corresponding author.}

% 单位
\affiliation[inst1]{organization={State Key Laboratory of Spatial Datum, College of Remote Sensing and Geoinformatics Engineering, Henan University},
  city={Zhengzhou}, postcode={450046}, country={China}}
\affiliation[inst2]{organization={School of Computer and Information Engineering, Henan University},
  city={Kaifeng}, postcode={475004}, country={China}}
\affiliation[inst3]{organization={School of Automation Science and Electrical Engineering, Beihang University},
  city={Beijing}, postcode={100091}, country={China}}

% 摘要与关键词（双栏模板会自动排在题头下；全文保持双栏）
\begin{abstract}
Deep Neural Networks (DNNs) are vulnerable to adversarial attacks, posing significant security threats to their deployment in remote sensing applications. Research on adversarial attacks not only reveals model vulnerabilities but also provides critical insights for enhancing robustness and resilience. Although current mixing-based strategies have been proposed to increase the transferability of adversarial examples, they either perform global blending or directly exchange a region in the images, which may destroy the global semantic features of the images and mislead the optimization direction of adversarial examples. Furthermore, their reliance on cross-entropy loss for perturbation optimization leads to gradient diminishing during iterative updates, directly compromising adversarial example quality. To address these limitations, we focus on non-targeted attacks and propose a novel framework via local mixing and logits optimization for generating transferable adversarial examples in remote sensing object recognition. First, we present a local mixing strategy to generate diverse yet semantically consistent inputs during the generation of adversarial examples.  Different from the typical MixUp method which globally blends two images and the MixCut method which stitches different images together, the proposed local mixing method merely blends local regions of two images to preserve as much of the global semantic information as possible. Second, we adapt the logit loss from target attacks to non-target attack scenarios, to mitigate the gradient vanishing problem inherent to cross entropy loss. Third, we have also applied a perturbation smoothing loss to suppress excessively high-frequency noise to further enhance the transferability of adversarial examples generated. Extensive experiments on FGSCR-42 and MTARSI datasets demonstrate superior performance over 12 state-of-the-art methods across 6 surrogate models. Notably, with ResNet as the surrogate model on MTARSI, our method achieves a 17.28\% average improvement in black-box attack success rate.
\end{abstract}


\maketitle

\section{Introduction}
With the rapid development of earth observation technology, a massive number of high-resolution remote sensing images are being collected from satellites and aerial platforms. These data play a crucial role in a wide range of applications, including natural disaster monitoring, urban planning, forest resource survey, and military reconnaissance. Effectively analyzing such vast quantities of remote sensing images has therefore become a pressing issue. In particular, remote sensing object recognition is to identify the type of an object in the image scenes, which provides essential support for higher-level applications such as content-based remote sensing image retrieval and remote sensing target detection.

Due to the powerful feature extraction ability, deep neural networks (DNNs) are widely applied to remote sensing object recognition \cite{59,28,14}. However, DNNs are vulnerable to adversarial attacks during real-world deployment. By crafting adversarial examples which involve subtle and nearly imperceptible changes to normal examples, attackers can easily mislead the DNN models to produce incorrect results \cite{5}. These attacks pose serious threats to system reliability and security, particularly in critical remote sensing applications where accurate interpretation is essential for decision-making.  Studying adversarial attacks on  DNN models for remote sensing object recognition is therefore crucial, as it not only reveals vulnerabilities but also provides critical insights for enhancing model robustness and resilience. 

According to the attacker’s knowledge of the target model, adversarial attacks can be categorized into white-box attacks \cite{7}, \cite{6}, \cite{8} and black-box attacks \cite{9}, \cite{10}, \cite{11}, \cite{12}, \cite{13}, \cite{14}, \cite{15}, \cite{16}, \cite{17}, \cite{18}.
White-box attacks refer to scenarios where the attackers have complete knowledge of the model’s architecture and parameters. The attackers can directly utilize these information to compute the model’s gradients and generate adversarial examples with strong attack capabilities. In contrast, black-box attacks assume that the attackers have no access to the internal information of the target model and can only infer the model’s behavior through input-output queries, making them more challenging. In real-world scenarios, black-box attacks are of greater practical significance because most models deployed in real environments do not expose their internal structures, and attackers usually cannot directly access the model parameters. According to the attacker's objective,  adversarial attacks can also be categorized into targeted \cite{58} \cite{47} and non-targeted attacks. Targeted attacks aim to mislead the model into classifying inputs as a specific class, while non-targeted attacks cause the model to misclassify inputs without any predefined target. We focus on the non-targeted black-box attacks in this study, which play a significant role in revealing model vulnerabilities.

To achieve black-box attacks, transferable attacks offer an effective approach. The basic idea is to craft adversarial examples on an accessible substitute model and leverage their transferability across different models so that they can also fool the target black-box model. When generating transferable adversarial examples, traditional transferable attack methods often suffer from overfitting to the substitute model and lead to degraded performance on other models. To address the overfitting problem, input transformation-based methods \cite{9}, \cite{10}, \cite{11}, \cite{12}, \cite{13}, \cite{14} have emerged in recent years, enhancing input diversity through data augmentation during  adversarial examples generation. By averaging the gradients over multiple transformed inputs, these methods mitigate overfitting and improve the transferability of the adversarial examples generated.

While existing research mainly focuses on adversarial attacks in the field of natural images, relevant findings indicate that such threats also exist in the field of remote sensing \cite{19}. For example,  to interfere with the models in the scene classification tasks \cite{20},  researchers have designed different attack methods through input transformations \cite{22}, multimodal data \cite{23}, and other techniques. However, existing attack methods fail to fully leverage the unique characteristics of remote sensing images. As shown in Figure 1, remote sensing images are characterized by highly complex and diverse backgrounds, ranging from urban landscapes to natural environments. Traditional input transformation methods augment individual images through operations like rotation, scaling, or cropping, essentially generating variations of the same scene content. This fundamentally limits adversarial example generalization across models. While several mixing-based strategies, such as MixUp/MixCut \cite{57}, and Admix \cite{35}, have been proposed to increase input diversity, they either perform global blending or directly exchange a region in the images, which may destroy the global semantic features of the images and reduce the quality of generated adversarial examples. Furthermore, these methods still use the typical cross entropy loss to optimize the perturbation, which suffers from gradient diminishing \cite{47} during iterative gradient updates. That is,  the loss gradient concerning the input diminishes toward zero as iterations increase. Since transferable attacks rely on substitute model gradients to update perturbations, this issue directly compromises adversarial example quality.

%Similarly,  physically realizable methods like complex texture patches \cite{24} and stripe attacks \cite{25} have been also proposed for object detection \cite{21}. 

\begin{figure}
    \centering
    \includegraphics[width=1\linewidth]{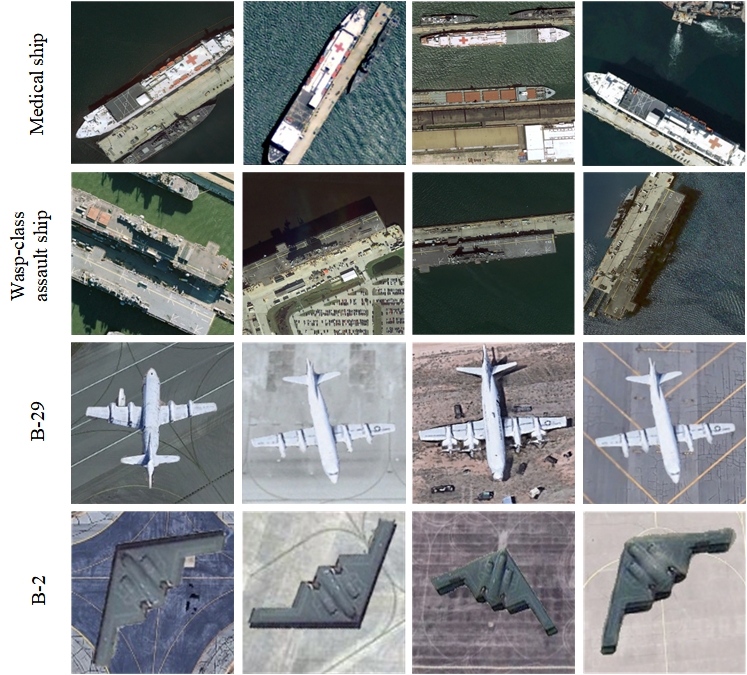}
    \caption{Remote sensing images from FGSCR-42 and MTARSI datasets demonstrate the complex background diversity and semantic consistency within each class category.}
    \label{fig:1}
\end{figure}

To address these limitations,  we in this paper focus on remote sensing object recognition task and propose a novel  method via local mixing and logits optimization for generating highly  transferrable adversarial examples. First, we present a local mixing strategy to generate diverse yet semantically consistent inputs during the generation of adversarial examples. Different from the typical MixUp method \cite{57}  which globally blends two images and the MixCut method \cite{57} which directly stitches different images together, the proposed local mixing method merely blends local regions of two images to preserve as much of the global semantic information as possible.  Second, we introduce a logits-based optimization specifically adapted for non-targeted attacks in object recognition. While previous work \cite{45} applied logit loss in targeted attack scenarios, we reformulate it for non-targeted attacks by directly minimizing the logits of the true class. Fundamentally different from traditional cross-entropy loss optimization, it bypasses the softmax operation that causes gradient vanishing \cite{44}, thereby maintaining stronger gradient signals throughout the iterative optimization process—a critical factor for transferable adversarial example generation. Third, acknowledging that excessively high-frequency perturbation components impair transferability\cite{17},  we have also applied a perturbation smoothing loss which uses a mean filter to suppress excessive high-frequency components to further enhance transferability across heterogeneous DNN architectures. The main contributions of this paper are as follows: 

1. We propose a local mixing strategy to alleviate overfitting in adversarial example generation for remote sensing object recognition. By merely blending the random regions of two images, our method leverages background diversity and generates diverse yet semantically consistent inputs.

2. We design a novel untargeted attack loss based on logits optimization. Unlike cross-entropy loss, our approach directly minimizes the true-class logits to avoid gradient vanishing, and integrates a perturbation-smoothing loss to suppress high-frequency noise, leading to more transferable adversarial examples across diverse architectures.

3. We conduct extensive experiments on two datasets, showing that our method consistently outperforms existing approaches. For example, on the MTARSI dataset with ResNet as the substitute model, we achieve up to 17.28\% average improvement in attack success rate.

The remainder of this paper is structured as follows. Section 2 briefly introduces the related work. Section 3 details the proposed method. Section 4 describes the datasets, the design and the results of the experiments. Finally, the conclusions are in section 5.

\section{Related work}
\subsection{Adversarial attacks}
Adversarial attacks are typically divided into two main categories: white-box attacks and black-box attacks. This classification is based on the extent to which the attacker has control over the target model \cite{29}. In white-box attacks, the attacker has full access to the target model’s architecture and parameters, allowing them to exploit the model's weaknesses to carefully design perturbations that achieve the desired attack outcome. Classic white-box attack methods include the Fast Gradient Sign Method (FGSM) \cite{5} and Projected Gradient Descent (PGD) \cite{30}. These methods optimize input perturbations using the gradient information of the target model, causing the model to make incorrect predictions.

In practical applications, since attackers typically cannot access detailed information about the target model, black-box attacks are more realistic and pose a significant threat to the application of deep neural networks. Currently, black-box attack methods are divided into two types: query-based attacks and transfer-based attacks. In query-based attacks \cite{31,32}, the attacker repeatedly sends query requests to the target model and generates adversarial examples based on the information provided by the model's feedback. However, due to the need for frequent query operations, these attacks are computationally expensive, particularly when handling large-scale datasets, where query costs can increase significantly. Transfer-based attacks are a viable black-box attack method, where adversarial examples are generated by attacking a substitute white-box model. These adversarial examples are then transferable and can effectively attack the black-box model as well. These classic white-box attack methods like FGSM \cite{5} and PGD \cite{30} do not perform well in transferability. Enhancing the transferability of adversarial examples remains a challenging problem. 

In recent years, researchers have proposed various optimization strategies to enhance the transferability of adversarial examples. Liu et al. \cite{16} introduced a method based on the ensemble of multiple substitute models, aiming to improve the transferability of adversarial examples across different models. Dong et al. \cite{15} proposed the Momentum Iterative Method (MIM), an iterative attack method that introduces momentum into the gradient updates to accelerate convergence and effectively enhance the transferability of adversarial examples. Gao et al. \cite{10} proposed PIM, which focuses not on the global image but on the patches of the input images, improving adversarial effectiveness by precisely interfering with these regions. Ding et al. \cite{18} explored a black-box attack method, OSFD, which interferes with intermediate layer features. By amplifying transformed features and separating them from the original features, OSFD improves the transferability of adversarial examples.

Input transformations, as a means of combating overfitting, have also been applied in recent years to improve transferability. Initially, input transformations were primarily focused on the spatial domain, where various transformations were applied to the input data. For example, Xie et al \cite{9} first proposed the Diversity Input Method (DIM), which enhanced the transferability of adversarial examples across different models by applying diverse random transformations to the input images. Guo et al. \cite{34} introduced BSR, which instead of transforming the entire input image, first divides the image into blocks, then performs random shuffling and rotation operations on these blocks. Later, Dong et al. \cite{11}  proposed TIM, which optimized the uniformity of input perturbations under spatial translations, making adversarial examples more transferable to target models. Lin et al. \cite{12} proposed SIM, which took advantage of the target model's invariance to input scale changes and enhanced the robustness and transferability of adversarial examples by simulating input images at different scales. Wang et al. \cite{35} introduced Admix, which linearly combines adversarial examples with other randomly selected natural images, simulating the diversity of input distributions and thereby enhancing transferability.% Liu et al.~\cite{58} proposed IDAA, a targeted attack method that applies data augmentations and spatial mixing to single images, employing targeted attack loss functions to generate adversarial examples. Their approach demonstrated effectiveness in targeted attack scenarios by leveraging spatial mixing strategies.

With the deepening of research, the application of input transformations has extended to the frequency domain. Zhang et al. \cite{13} proposed SSA, which aims to enhance the diversity of adversarial examples in the frequency domain through spectral transformations, thereby generating more transferable adversarial examples that can effectively attack various defense models. Unlike natural images, in the field of remote sensing target recognition, researchers have begun to explore more complex transformation strategies by combining the characteristics of remote sensing data. Fu et al. \cite{14} discovered that the key to remote sensing object recognition lies in the features of the target-related regions. To address this, they proposed the SFCoT method, which combines frequency and spatial domain transformations to achieve effective decoupling of the target and background. Subsequently, the method applies block-wise input transformations to the decoupled regions, significantly enhancing the transferability of the model.

\subsection{Adversarial defense}
Adversarial defense aims to enhance the ability of deep learning models to resist adversarial examples attacks. Existing defense strategies primarily include model optimization, adversarial training, and data preprocessing methods.  

In model optimization approaches, the main idea is to improve the robustness of models by modifying the training process or regularizing the model's parameters to better resist adversarial perturbations. Papernot et al. \cite{36} proposed defensive distillation, training models via a distillation algorithm using output probability distributions as new labels. Addepalli et al. \cite{37} introduced a novel Bit Plane Feature Consistency (BPFC) regularization method, improving adversarial robustness by constraining consistency across different quantized image representations. Additionally, Liao et al. \cite{38} proposed a high-level representation-guided denoiser, using a U-Net denoising network to remove noise from images and counter adversarial attacks.  

Adversarial training was first proposed by Goodfellow et al. \cite{5}, utilizing the Fast Gradient Sign Method (FGSM) to generate adversarial examples, which are then mixed with original images for training. Madry et al. \cite{30} improved adversarial training by introducing a Projected Gradient Descent (PGD)-based method. Tramer et al. \cite{39} proposed ensemble adversarial training, using adversarial examples generated from multiple pretrained models to enhance robustness against cross-model adversarial examples. Kannan et al. \cite{40} introduced a logit pairing strategy based on PGD adversarial training, proposing a hybrid mini-batch PGD method. By pairing clean images with adversarial or other clean ones, this approach significantly improves model robustness under both white-box and black-box attacks.  

In data preprocessing-based adversarial defense methods, the core idea is to modify the input data in a way that reduces the effect of adversarial perturbations before they reach the model. Guo et al. \cite{41} proposed input transformation techniques such as image scaling, cropping, and compression to eliminate perturbations while preserving useful information. Xie et al. \cite{42} suggested random resizing and padding of images. They also proposed a pixel deflection-based method, perturbing pixel values locally and combining wavelet denoising techniques to effectively reduce damage from pixel shifts and adversarial perturbations. Jin et al. \cite{43} introduced the APE-GAN strategy, utilizing a generator to simultaneously reconstruct adversarial and clean examples.
\begin{figure*}[htbp]
    \centering
    \includegraphics[width=\linewidth]{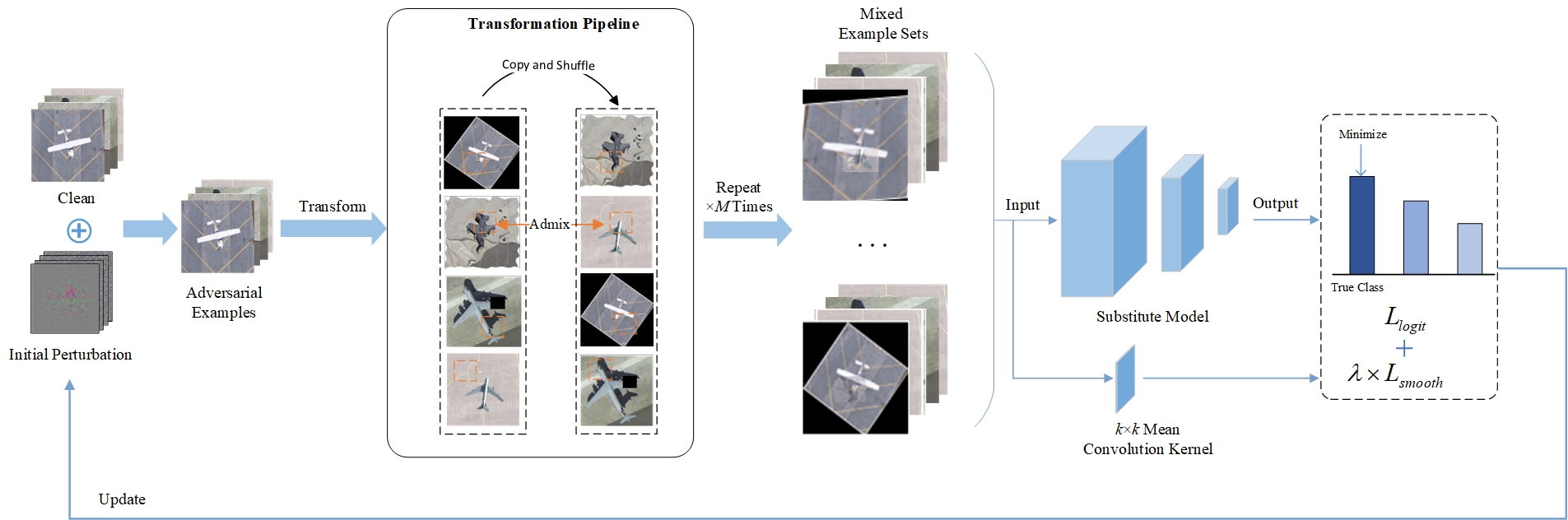}
    \caption{Overview of our proposed method. Adversarial examples are first passed through a transformation pipeline to introduce diversity. Mixed images are passed through a surrogate model to compute logit loss (minimizing true class logit) and smoothing loss via a $k \times k$ mean convolution kernel. We then compute the smooth gradient and combine it with momentum to iteratively update the perturbation.}
    \label{fig:2}
\end{figure*}

\section{Methodology}
\subsection{Problem description}
Given a target model \( f \), a clean image \( x \), and its true label \( y \), an adversarial attack aims to construct an adversarial example \( x^{adv} \) to deceive the model such that:
\begin{equation}
f(x^{adv}) \neq f(x) = y
\label{eq:adv_example}
\end{equation}

where \( x^{adv} \) is generated by adding a small perturbation \( \delta \) to \( x \), and this perturbation is visually imperceptible to human observers, satisfying
\begin{equation}
\| x^{adv} - x \|_p \leq \varepsilon
\label{eq:adv_distance}
\end{equation}

where \( \| \cdot \|_p \) denotes the \( p \)-norm. Following previous works \cite{15}, \cite{9}, \cite{11}, \cite{12}, \cite{13}, \cite{14}, we adopt the \( \infty \)-norm (\( p = \infty \)). We formulate this optimization problem through Eq. 1, where \( L \) is the loss function and \( \varepsilon \) controls the magnitude of the perturbation
\begin{equation}
x^{adv} = \arg\max L, \quad \text{s.t.} \quad \| x^{adv} - x \|_\infty \leq \varepsilon
\label{eq:adv_max}
\end{equation}

\subsection{The overview of the proposed method}
To address this optimization challenge, we propose a lightweight framework via local mixing and logits Optimization for remote sensing object recognition task, to generate adversarial examples with superior transferability.

Fig.~\ref{fig:2} comprehensively illustrates the workflow of our method. Initially, It begins with a batch of clean images and a adversarial perturbation. After adding the clean images with the adversarial perturbation to generate the adversarial examples,  a data augmentation operation is randomly selected from Table.~\ref{tab:DataAugmentation} and applied to the adversarial examples, creating a set of augmented images. These augmentation operations—including rotation, scaling, and cropping—effectively simulate various real-world image variations. Subsequently, it duplicates this set of augmented images and randomly shuffles their order to create potential mixing candidates. It then randomly selects rectangular regions within the images and apply the Admix technique to proportionally blend the selected regions of the original set of augmented images with the regions from their shuffled counterparts, generating a mixed image set. Notably, it repeats above transformation process multiple times, each time applying different mixing regions and data augmentation operations. The resulting sets of mixed images and their corresponding labels are concatenated across these diverse transformations and used jointly for gradient computation. This strategy effectively explores a broader perturbation space, significantly enhancing the transferability of the final adversarial examples.

To optimize perturbation search, it develops a composite loss function that combines logit loss with perturbation smoothing loss. Based on the MI-FGSM framework, it utilizes momentum terms to accumulate historical gradient information and reduce optimization noise and fluctuations. Additionally, it incorporates gradient smoothing to reduce overfitting and improve the transferability of adversarial examples. This combination of strategies encourages smoother gradients and more stable update directions, leading to adversarial examples with improved transferability in black-box settings.

The entire algorithm workflow is summarized in Algorithm.~\ref{alg:adversarial}, which clearly demonstrates the complete process from initialization to final adversarial example generation, including key components such as input transformation, gradient calculation, momentum updates, and adversarial updates during the iteration process. We detail the key components of the proposed method as follows.

\subsection{Local mixing based input transformation}
In adversarial example generation, data augmentation operations are typically applied to input examples to enhance input diversity and reduce overfitting to the substitute model. However, while enhancing input diversity, these operations may also potentially damage the semantic information of the samples. In extreme cases, a  data augmentation operation could destroy key semantic features of the sample and  leads to recognition errors in the model. Although such errors are expected by adversarial samples, they are not caused by the perturbation we add. Using the loss generated by this error to optimize the perturbation will lead to deviations in the optimization direction. Therefore, preserving the semantic information of the samples as much as possible during data augmentation is crucial for obtaining high-quality perturbation. To achieve this purpose, we propose a local mixing based input transformation strategy, whose pipeline is illustrated in Fig.~\ref{fig:2}.

For a batch of images $x = {x_1, x_2, \ldots, x_b}$ which are added by a perturbation, it first applies data augmentation to each image in the batch, generating a transformed image set
\begin{equation}
S_1 = \{ T(x_1), T(x_2), \dots, T(x_b) \}
\label{eq:S1}
\end{equation}
where $T(\cdot)$ represents the data augmentation function, which randomly selects one of the transformation operations which are shown in Table~\ref{tab:DataAugmentation}. These transformations simulate various changes of images in the real world, helping the generated adversarial examples achieve stronger generalization ability.

To further increase the diversity of inputs, we duplicate the augmented image set and apply random shuffling.
\begin{equation}
S_2 = \text{shuffle}(S_1)
\label{eq:S2}
\end{equation}

where \( \text{shuffle}(\cdot) \) represents the random permutation function. We then adopt a local region-mixing strategy. For each sample index $i$ (where $i = 1, \dots, b$), we randomly select a rectangular region and denote its binary mask as $R_i$. Within this rectangular region, we blend $S_1[i]$ and $S_2[i]$ using mixing weight $\eta \in (0,1)$, while keeping the areas outside the rectangular region unchanged from $S_1[i]$.

\begin{equation}
\tilde{x}_i = R_i \odot \big( \eta\, S_1[i] + (1-\eta)\, S_2[i] \big)
            + (1 - R_i) \odot S_1[i]
\label{eq:R}
\end{equation}

By shuffling the order of images in $S_2$, we break the fixed correspondence with $S_1$, thereby achieving mixing effects between different images to enhance diversity.

\subsection{Smoothness-Regularized logit loss}
The design of the loss function is crucial for generating effective adversarial perturbations, as it directly determines the quality and stability of the gradients used for optimizing the perturbations. When using the typical cross-entropy loss to optimize the perturbation, the softmax function compresses the logit vector containing arbitrary real numbers into a probability distribution, where each element's value ranges between 0 and 1, and the sum of all elements is 1. When there is one or more extremely large (positive or negative) values in the logit vector, the softmax function enters a saturated state, and its output tends towards a deterministic distribution, leading to an extremely small gradient. This makes it difficult to further optimize the perturbation. Meanwhile, there are usually high-frequency noise in the perturbation. Because different neural networks exhibit varying sensitivities to high-frequency components in the input space, high-frequency noise may drive the attack to exploit model-specific patterns during perturbation optimization. This will harm the transferability of the adversarial examples across different models. To address these issues, we propose a combined loss function that integrates logit loss and perturbation smoothing loss to mitigate gradient vanishing and explicitly penalize high-frequency components. The definitions and roles of these two loss functions are detailed below.

Given an input \( x \) and an adversarial perturbation \( \delta \), the logit output of the target model \( f \) is
\begin{equation}
z = f(x + \delta)
\label{eq:z}
\end{equation}
where \( z \in \mathbb{R}^C \) represents the unnormalized output before softmax (with \( C \) being the number of classes). Let \( z_y \) denote the logit value of the model for the true class \( y \). The logit loss is defined as
\begin{equation}
L_{logit} = -z_y
\label{eq:Logit}
\end{equation}

In targeted attack scenarios, the common approach is to maximize the logit value of the target class to strengthen the model’s prediction toward that class. In the untargeted attack scenario of this paper, the proposed method minimizes the logit of the true class to weaken the model’s confidence in the correct prediction, making misclassification more likely. 

Moreover, to prevent the excessive amplification of high-frequency noise during iterative optimization, the proposed method introduces a smoothing loss to smooth the adversarial perturbation and constrain its frequency distribution. Specifically,  a low-pass filter is applied to the perturbation \( \delta \) and  the filtered result is taken as a regularization term to limit high-frequency components
\begin{equation}
L_{smooth} = \|\text{LowPassFilter}(x^{adv}-x) \|_1
\label{eq:low-pass}
\end{equation}
where \( \text{LowPassFilter} \) denotes a \( k \times k \) mean convolution kernel that attenuates high-frequency components by averaging pixel values within a sliding window. 

Ultimately, the logit loss and smoothing loss are combined with specific weights to form a comprehensive loss function.
\begin{equation}
L_{total} = L_{logits} + \lambda L_{smooth}
\label{eq:Total}
\end{equation}
where \( \lambda \) is a hyperparameter that balances the contributions of the loss terms.

\begin{table}[h!]
\centering
\caption{Data Augmentation Strategies in Our Method} 
\setlength{\tabcolsep}{1pt} % Reduce horizontal spacing between columns
\begin{tabular}{ccccc}
\toprule[1pt]
    Origin & HFlip & VFlip & Rotate & Perspective \\ 
    \midrule
    \\[-1.9ex]
    \includegraphics[width=0.18\columnwidth]{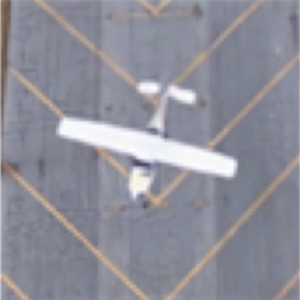} & 
    \includegraphics[width=0.18\columnwidth]{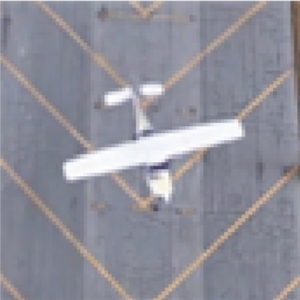} & 
    \includegraphics[width=0.18\columnwidth]{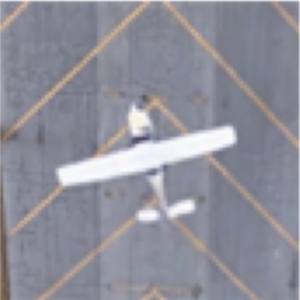} & 
    \includegraphics[width=0.18\columnwidth]{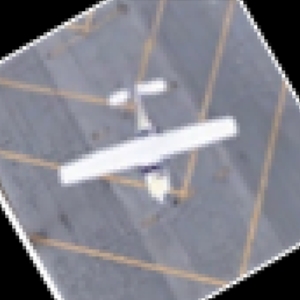} & 
    \includegraphics[width=0.18\columnwidth]{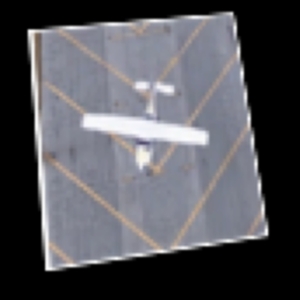} \\ 
    \midrule
    Affine & Elastic & Spline & Resize & Erasing \\ 
    \midrule
    \\[-1.9ex]
    \includegraphics[width=0.18\columnwidth]{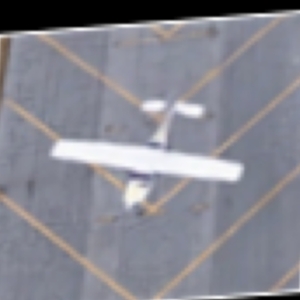} & 
    \includegraphics[width=0.18\columnwidth]{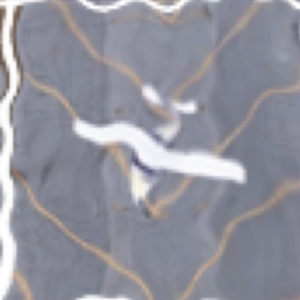} & 
    \includegraphics[width=0.18\columnwidth]{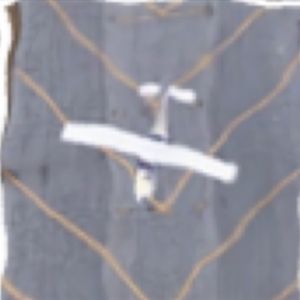} & 
    \includegraphics[width=0.18\columnwidth]{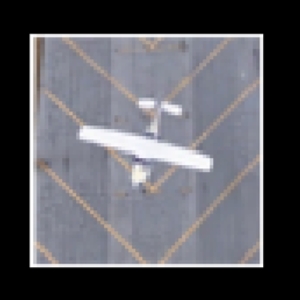} & 
    \includegraphics[width=0.18\columnwidth]{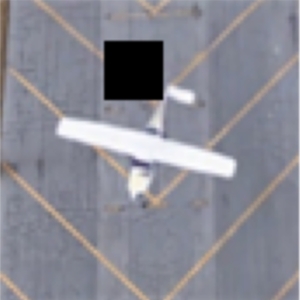} \\ 
    \bottomrule[1pt]
\end{tabular}
\label{tab:DataAugmentation}
\end{table}

\subsection{Gradient calculation and perturbations optimization}
To obtain smooth gradients and stable update directions, we repeat the above local mixing based input transformation $M$ times, where each transformation generates a batch of images for gradient calculation. Multiple rounds of transformation compensate for the limitations of a single transformation, making the gradient more stable and with stronger generalization capability. 

As shown in the formula Eq.~\ref{eq:g}, we calculate the gradients for each of these $M$ batches and average them to obtain the gradient $\overline{g}_{t+1}$. Based on the gradient, we further adopt a gradient smoothing strategy as shown in Eq.~\ref{eq:g2} to generate a momentum term to accelerate convergence.

\begin{equation}
\bar{g}_{t+1} = \frac{1}{M}\sum_{m=1}^{M}\nabla_{x_t^{adv}}L_{total}(\tilde{x}_t^{(m)}, y) 
\label{eq:g}
\end{equation}

\begin{equation}
g_{t+1} = g_t + \frac{\bar{g}_{t+1}}{\|\bar{g}_{t+1}\|_1} 
\label{eq:g2}
\end{equation}

After computing the gradient, the adversarial example update process is designed as Eq.~\ref{eq:g4}. 

\begin{equation}
x_{t+1}^{adv} = Clip_{x,\varepsilon}\{x_t^{adv} + \alpha \cdot sign(g_{t+1})\}
\label{eq:g4}
\end{equation}

To ensure the adversarial perturbations remain imperceptible, we constrain them within an $L_\infty$ ball of radius $\varepsilon$ centered at the clean image $x$. As shown in Eq.~\ref{eq:g4}, each PGD iteration updates the adversarial example and then applies the clipping operation $\mathrm{Clip}_{x,\varepsilon}$ to project $x^{adv}_t$ back into the feasible region $[x-\varepsilon,\,x+\varepsilon]$.
% needed in second column of first page if using \IEEEpubid
%\IEEEpubidadjcol

\begin{algorithm}
\caption{Adversarial Example Generation Algorithm}
\label{alg:adversarial}
\begin{algorithmic}[1]
\State \textbf{Input:} A substitute model $f$, loss function $L$, a batch of input images $x$ with label $y$, supplement $\alpha$, perturbation budget $\varepsilon$, transformation times $M$, number of iterations $T$, and mixing strength $\eta$.
\State \textbf{Output:} A batch of adversarial examples.
\State $g_0 \gets 0$, $x_0^{\text{adv}} \gets x$
\For{$t \gets 0$ \textbf{to} $T-1$}
    \For{$m \gets 1$ \textbf{to} $M$}
        \State Construct a transformed image set and concatenate.
    \EndFor
    \State Calculate the smooth gradient $\bar{g}_{t+1}$ of the transformed image set by Eq.~\ref{eq:g}.
    \State Update the momentum by Eq.~\ref{eq:g2}.
    \State Update the generated adversarial example $x_{t+1}^{\text{adv}}$ based on the sign of ${g}_{t+1}$.
\EndFor
\State \textbf{return} $x^{\text{adv}}$
\label{al:1}
\end{algorithmic}
\end{algorithm}

\section{Experiments}

\subsection{DataSets}
The proposed method is tested on the FGSCR-42 \cite{49} and MTARSI \cite{50} datasets.
FGSCR-42 is a publicly available dataset specifically designed for fine-grained classification of ships in remote sensing images. It contains 7,776 high-definition remote sensing images, covering 42 different ship categories, including military ships (e.g., aircraft carriers, destroyers) and civilian ships (e.g., cargo ships, cruise ships). These images are primarily collected from Google Earth. 

MTARSI is a benchmark dataset for aircraft type recognition in remote sensing images, containing 9,589 high-definition remote sensing images covering 20 aircraft types. These images are collected from Google Earth and have complex backgrounds, varying spatial resolutions, and differences in pose, location, illumination, and time periods.

Fig.~\ref{fig:1} shows some images from the FGSCR-42 and MTARSI datasets. We resize all images to 224×224 pixels. For each class, we randomly divide the dataset into training and testing sets in a 4:1 ratio. After training our model on the training set, we conduct adversarial attack experiments on the test set. The FGSCR-42 dataset can be downloaded from \url{https://github.com/DYH666/FGSCR-42}, and MTARSI can be downloaded from \url{https://www.kaggle.com/datasets/aqibriaz/mtarsidataset}.

\subsection{Model architectures}
Using a variety of models with different architectures can help researchers analyze the transferability of adversarial attacks. In this study, we carefully selected six classic deep neural network models which have been widely used in related works to evaluate the effectiveness of adversarial attacks. They  are  VGG-16\cite{53}, VGG-19\cite{53}, ResNet-34\cite{54}, ResNet-50\cite{54}, DenseNet-121\cite{55}, and Inception-ResNet-V2\cite{56}. From the relatively shallow VGG to the very deep DenseNet/Inception-ResNet, these models have demonstrated outstanding performance in object recognition and represent key milestones in the development of deep learning. VGG-16 and VGG-19 increase the network depth by stacking multiple 3x3 convolutional layers, emphasizing the importance of depth for performance improvement. ResNet-34 and ResNet-50 introduced residual connections, overcoming the bottleneck of training deep networks. DenseNet-121 employs dense connections to optimize feature reuse and information flow. Inception-ResNet-V2 combines the multi-scale feature extraction of the inception module and the residual connections of ResNet, showcasing the unique advantages of heterogeneous architectures. 

By considering a variety of classic models, we ensure that the results of the adversarial attack analysis are not limited to a single architecture. This is of significant importance for developing universal adversarial defense strategies and improving the overall security of deep learning models.

\subsection{Comparison methods}
To validate the performance of the proposed method, we compared it with several advanced attack methods, including PGD \cite{8}, MIM \cite{15}, TIM \cite{11}, PIM \cite{10}, DIM \cite{9}, SIM \cite{12}, Admix \cite{35}, Mixup/MixCut \cite{57}, SSA \cite{13}, BSR \cite{34}, and SFCoT \cite{14}. Specifically, PGD represents the standard iterative attack based on basic gradient updates. MIM introduces momentum optimization to stabilize and guide the update direction. TIM enhances transferability by applying translation-invariant smoothing. PIM focuses on pixel-level attention to refine perturbations. DIM improves robustness through diverse input transformations. SIM leverages scale-invariant transformations to boost attack generalization. Admix, Mixup, and Mixcut are data mixing strategies that generate new adversarial examples by combining multiple inputs. SSA and BSR further explore the spatial and spectral domains to craft more transferable perturbations. Additionally, SFCoT decouples target and background features in the frequency domain for targeted perturbation optimization. 

\subsection{Implementation details}
In our experiments, the target models were trained on the training set, and adversarial attacks were conducted on the test set samples that all models classified correctly. During the generation of adversarial examples, the perturbation constraint \( \varepsilon \), step size \( \alpha \), and number of iterations \( T \) were set to 16, 1, and 30, respectively. The local mixing based input transformation times M used for gradient computation were set to 25, and the loss weight \( \lambda \) was set to 200. These parameters (\( \varepsilon \), \( \alpha \), \( T \)) were kept consistent with the settings in SFCoT. In the input transformation phase,  the mixing strength $\eta$ was set to 0.5. Parameters specific to each comparison method followed their recommended configurations.  Experiments were conducted in the PyTorch framework with support from an Nvidia RTX 4090 GPU.

%the number of gradient averaging \( N \) was set to 5, with

\subsection{Performance comparison}
\subsubsection*{1) Quantitative Results}
To validate the effectiveness of the proposed method, we compared its performance with selected mainstream adversarial attack methods across multiple model architectures, using attack success rate(ASR) as the evaluation metric. ASR refers to the proportion of images that successfully induce the model to make incorrect predictions among all attacked images. It is calculated as follows 
\begin{equation}
\text{ASR} = \frac{N_{\text{error}}}{N_{\text{total}}} \times 100\%
\label{eq:asr}
\end{equation}
where \( N_{\text{error}} \) denotes the number of successful adversarial examples, and \( N_{\text{total}} \) denotes the total number of adversarial examples. A higher ASR indicates a more effective adversarial attack. 

The comparison results on two datasets are presented in Table~\ref{longtab:1} and Table~\ref{longtab:2}. In these tables, the first column represents the substitute model, the second column lists mainstream attack methods, and each subsequent column indicates the attack success rate of adversarial examples generated on the substitute model against the corresponding target model. For the MTARSI dataset, due to identical dataset splits, we directly used the experimental data from SFCoT \cite{14}. For the FGSCR-42 dataset, due to differences in dataset splits, we re-conducted experiments for all comparison methods using their recommended configurations to ensure fairness.

The results demonstrate that our method significantly outperforms state-of-the-art approaches across different models. On the FGSCR-42 dataset (Table~\ref{longtab:1}), Compared with the best performance among all comparison methods, our approach achieves an average improvement of 8.84\% and a maximum improvement of 25.81\% in black-box ASR. In same-architecture attacks between VGG-16 and VGG-19, and between ResNet-18 and ResNet-50, we achieved a 100\% ASR. When using Inception-ResNet-v2 as a surrogate model, the highest average black-box ASR was achieved, reaching 14.92\%. When using DenseNet-121 as a surrogate model to attack the VGG19 architecture, we achieved the highest ASR improvement, reaching 25.81\%. On the MTARSI dataset (Table~\ref{longtab:2}), compared with the best performance among all comparison methods, our approach achieves an average improvement of 12.15\% and a maximum improvement of 32.28\% in black-box ASR.  When using ResNet-18 and ResNet-50 architectures as surrogate models, we achieved the highest average black-box ASR improvements compared to all comparison methods, with improvements of 17.14\% and 17.43\% respectively. Particularly, when ResNet-50 attacks Inception-ResNet-v2, it achieved a 32.28\% improvement compared to state-of-the-art performance. Overall, the results demonstrate the superior performance of our method against heterogeneous architectures, consistently outperforming state-of-the-art techniques. Table~\ref{tab:image_comparison} shows the adversarial examples generated by different attack methods, which have minimal visual differences compared to the original examples.

\begin{table*}[t]
\centering
\footnotesize
\renewcommand{\arraystretch}{0.8}
\caption{Attack Success Rates (\%) on FGSCR-42 Dataset, ↑ indicates performance improvement over the best baseline.}

\begin{tabular}{@{}>{\centering\arraybackslash}p{1.5cm}|>{\centering\arraybackslash}p{1.5cm}|>{\raggedright\arraybackslash}p{2cm}>{\raggedright\arraybackslash}p{2cm}>{\raggedright\arraybackslash}p{2cm}>{\raggedright\arraybackslash}p{2cm}>{\raggedright\arraybackslash}p{2cm}>{\raggedright\arraybackslash}p{2cm}}
\toprule
Model & Attack & Vgg-16 & Vgg-19 & Res-34 & Res-50 & Dense-121 & IncRes-v2 \\
\midrule

\multirow{13}{*}{Vgg-16} 
  & PGD    & 100.00 & 99.25  & 75.77 & 87.26 & 78.85 & 38.13 \\
  & MIM    & 99.25  & 98.66  & 78.93 & 87.34 & 82.26 & 49.70 \\
  & PIM    & 100.00 & 99.16  & 75.52 & 87.17 & 81.84 & 41.38 \\
  & TIM    & 100.00 & 99.83  & 67.02 & 71.19 & 69.85 & 30.72 \\
  & DIM    & 100.00 & 100.00 & 94.69 & 91.42 & 89.34 & 36.30 \\
  & SIM    & 100.00 & 100.00 & 82.68 & 92.17 & 90.09 & 37.63 \\
  & Admix  & 100.00 & 99.50  & 82.09 & 91.09 & 87.51 & 51.54 \\
  & Mixup  & 90.59  & 90.59  & 82.09 & 87.76 & 82.01 & 52.45 \\
  & Mixcut & 100.00 & 100.00 & 88.75 & 84.83 & 93.67 & 53.20 \\
  & SSA    & 99.16  & 97.83  & 76.51 & 83.93 & 80.01 & 48.62 \\
  & BSR    & 99.91  & 99.83  & 87.09 & 93.33 & 92.08 & 48.04 \\
  & SFCoT  & 100.00 & 99.91  & 94.50 & 97.08 & 95.17 & 52.20 \\
  & \cellcolor[gray]{0.9}Ours   
  & \cellcolor[gray]{0.9}\textbf{100.00} {\footnotesize\textcolor{darkergreen}{(0.00↑)}}
  & \cellcolor[gray]{0.9}\textbf{100.00} {\footnotesize\textcolor{darkergreen}{(0.00↑)}} 
  & \cellcolor[gray]{0.9}\textbf{99.33} {\footnotesize\textcolor{darkergreen}{(4.64↑)}} 
  & \cellcolor[gray]{0.9}\textbf{99.33} {\footnotesize\textcolor{darkergreen}{(2.25↑)}} 
  & \cellcolor[gray]{0.9}\textbf{98.33} {\footnotesize\textcolor{darkergreen}{(3.16↑)}} 
  & \cellcolor[gray]{0.9}\textbf{60.36} {\footnotesize\textcolor{darkergreen}{(7.16↑)}} \\
\midrule

\multirow{13}{*}{Vgg-19} 
  & PGD    & 97.16  & 100.00 & 71.27 & 85.17 & 80.51 & 36.55 \\
  & MIM    & 97.50  & 99.08  & 74.85 & 86.01 & 82.26 & 48.62 \\
  & PIM    & 97.83  & 100.00 & 74.68 & 86.76 & 81.76 & 40.21 \\
  & TIM    & 92.58  & 100.00 & 66.44 & 68.02 & 68.44 & 31.05 \\
  & DIM    & 99.50  & 100.00 & 82.26 & 90.50 & 89.17 & 37.13 \\
  & SIM    & 99.33  & 100.00 & 81.68 & 89.09 & 88.25 & 36.80 \\
  & Admix  & 98.91  & 100.00 & 82.18 & 89.67 & 86.92 & 50.70 \\
  & Mixup  & 92.50  & 90.75  & 79.93 & 87.76 & 84.09 & 51.70 \\
  & Mixcut & 99.66  & 100.00 & 88.00 & 92.58 & 91.42 & 51.20 \\
  & SSA    & 96.31  & 90.16  & 72.77 & 84.09 & 79.51 & 46.71 \\
  & BSR    & 99.41  & 100.00 & 83.68 & 91.09 & 90.92 & 45.71 \\
  & SFCoT  & 99.83  & 100.00 & 89.17 & 94.92 & 94.25 & 48.62 \\
  & \cellcolor[gray]{0.9}Ours   
  & \cellcolor[gray]{0.9}\textbf{100.00} {\footnotesize\textcolor{darkergreen}{(0.17↑)}} 
  & \cellcolor[gray]{0.9}\textbf{100.00} {\footnotesize\textcolor{darkergreen}{(0.00↑)}} 
  & \cellcolor[gray]{0.9}\textbf{97.58} {\footnotesize\textcolor{darkergreen}{(8.41↑)}} 
  & \cellcolor[gray]{0.9}\textbf{98.58} {\footnotesize\textcolor{darkergreen}{(3.66↑)}} 
  & \cellcolor[gray]{0.9}\textbf{98.33} {\footnotesize\textcolor{darkergreen}{(4.08↑)}} 
  & \cellcolor[gray]{0.9}\textbf{53.03} {\footnotesize\textcolor{darkergreen}{(1.33↑)}} \\
\midrule

\multirow{13}{*}{Res-34} 
  & PGD    & 18.23  & 11.15  & 100.00 & 68.44 & 49.04 & 32.47 \\
  & MIM    & 56.28  & 35.97  & 100.00 & 78.01 & 66.77 & 53.78 \\
  & PIM    & 39.38  & 21.89  & 100.00 & 71.94 & 60.03 & 39.05 \\
  & TIM    & 16.31  & 7.16   & 100.00 & 43.37 & 88.80 & 17.65 \\
  & DIM    & 44.79  & 30.39  & 100.00 & 83.68 & 75.27 & 27.56 \\
  & SIM    & 43.79  & 28.97  & 100.00 & 84.76 & 75.60 & 27.89 \\
  & Admix  & 74.68  & 62.53  & 100.00 & 88.42 & 85.67 & 55.62 \\
  & Mixup  & 77.10  & 64.44  & 98.08  & 88.84 & 84.42 & 54.45 \\
  & Mixcut & 79.01  & 67.44  & 100.00 & 93.08 & 90.34 & 55.95 \\
  & SSA    & 67.77  & 57.45  & 100.00 & 93.42 & 84.42 & 51.70 \\
  & BSR    & 80.34  & 69.27  & 100.00 & 97.16 & 94.17 & 54.62 \\
  & SFCoT  & 88.42  & 79.01  & 100.00 & 99.41 & 97.75 & 55.70 \\
  & \cellcolor[gray]{0.9}Ours   
  & \cellcolor[gray]{0.9}\textbf{97.83} {\footnotesize\textcolor{darkergreen}{(9.41↑)}} 
  & \cellcolor[gray]{0.9}\textbf{95.50} {\footnotesize\textcolor{darkergreen}{(16.49↑)}} 
  & \cellcolor[gray]{0.9}\textbf{100.00} {\footnotesize\textcolor{darkergreen}{(0.00↑)}} 
  & \cellcolor[gray]{0.9}\textbf{100.00} {\footnotesize\textcolor{darkergreen}{(0.59↑)}} 
  & \cellcolor[gray]{0.9}\textbf{99.58} {\footnotesize\textcolor{darkergreen}{(1.83↑)}} 
  & \cellcolor[gray]{0.9}\textbf{68.44} {\footnotesize\textcolor{darkergreen}{(12.49↑)}} \\
\midrule

\multirow{13}{*}{Res-50} 
  & PGD    & 17.23  & 10.15  & 57.86  & 100.00 & 53.37 & 32.30 \\
  & MIM    & 51.45  & 32.80  & 72.10  & 100.00 & 71.35 & 53.78 \\
  & PIM    & 33.88  & 19.73  & 69.60  & 100.00 & 67.61 & 37.80 \\
  & TIM    & 15.65  & 7.74   & 45.87  & 100.00 & 45.87 & 16.98 \\
  & DIM    & 41.38  & 30.72  & 85.76  & 100.00 & 90.09 & 28.72 \\
  & SIM    & 42.71  & 30.55  & 94.09  & 100.00 & 90.75 & 28.89 \\
  & Admix  & 72.68  & 60.86  & 91.09  & 100.00 & 91.50 & 54.62 \\
  & Mixup  & 74.52  & 62.61  & 88.84  & 97.58  & 88.09 & 55.87 \\
  & Mixcut & 73.18  & 62.94  & 94.00  & 100.00 & 94.25 & 55.53 \\
  & SSA    & 68.10  & 58.86  & 90.25  & 100.00 & 89.67 & 50.70 \\
  & BSR    & 73.43  & 62.44  & 95.75  & 100.00 & 98.91 & 53.37 \\
  & SFCoT  & 81.01  & 73.68  & 98.50  & 100.00 & 99.16 & 54.53 \\
  & \cellcolor[gray]{0.9}Ours   
  & \cellcolor[gray]{0.9}\textbf{97.58} {\footnotesize\textcolor{darkergreen}{(16.57↑)}} 
  & \cellcolor[gray]{0.9}\textbf{96.50} {\footnotesize\textcolor{darkergreen}{(22.82↑)}} 
  & \cellcolor[gray]{0.9}\textbf{100.00} {\footnotesize\textcolor{darkergreen}{(1.50↑)}} 
  & \cellcolor[gray]{0.9}\textbf{100.00} {\footnotesize\textcolor{darkergreen}{(0.00↑)}} 
  & \cellcolor[gray]{0.9}\textbf{100.00} {\footnotesize\textcolor{darkergreen}{(0.84↑)}} 
  & \cellcolor[gray]{0.9}\textbf{68.85} {\footnotesize\textcolor{darkergreen}{(12.98↑)}} \\
\midrule

\multirow{13}{*}{\parbox{1.5cm}{\centering Dense-121}}    
    & PGD    & 21.23  & 14.32  & 56.53  & 72.43  & 100.00 & 32.72 \\   
    & MIM    & 57.86  & 42.96  & 72.18  & 83.76  & 100.00 & 53.28 \\   
    & PIM    & 41.38  & 28.80  & 71.94  & 82.68  & 100.00 & 40.21 \\   
    & TIM    & 18.15  & 8.74   & 46.71  & 51.12  & 100.00 & 18.23 \\   
    & DIM    & 46.79  & 34.30  & 81.84  & 92.00  & 100.00 & 27.81 \\   
    & SIM    & 45.96  & 34.97  & 80.59  & 91.75  & 100.00 & 28.97 \\   
    & Admix  & 76.18  & 65.69  & 86.34  & 89.75  & 100.00 & 54.28 \\   
    & Mixup  & 78.68  & 65.44  & 84.92  & 87.59  & 97.41  & 54.03 \\   
    & Mixcut & 78.01  & 66.27  & 88.92  & 92.58  & 100.00 & 54.78 \\   
    & SSA    & 66.11  & 60.11  & 81.59  & 89.84  & 100.00 & 50.54 \\   
    & BSR    & 74.00  & 63.28  & 90.42  & 94.92  & 100.00 & 52.87 \\   
    & SFCoT  & 81.34  & 70.94  & 95.50  & 99.80  & 100.00 & 53.03 \\   
    & \cellcolor[gray]{0.9}Ours   
    & \cellcolor[gray]{0.9}\textbf{96.75} {\footnotesize\textcolor{darkergreen}{(15.41↑)}} 
    & \cellcolor[gray]{0.9}\textbf{96.75} {\footnotesize\textcolor{darkergreen}{(25.81↑)}} 
    & \cellcolor[gray]{0.9}\textbf{99.66} {\footnotesize\textcolor{darkergreen}{(4.16↑)}} 
    & \cellcolor[gray]{0.9}\textbf{99.83} {\footnotesize\textcolor{darkergreen}{(0.03↑)}} 
    & \cellcolor[gray]{0.9}\textbf{100.00} {\footnotesize\textcolor{darkergreen}{(0.00↑)}} 
    & \cellcolor[gray]{0.9}\textbf{69.60} {\footnotesize\textcolor{darkergreen}{(14.82↑)}} \\
\midrule

\multirow{13}{*}{\parbox{1.5cm}{\centering IncRes-v2}}    
& PGD    & 4.99   & 3.49   & 16.31  & 29.05  & 20.23  & 100.00 \\   
& MIM    & 11.90  & 7.32   & 24.97  & 37.30  & 32.22  & 100.00 \\   
& PIM    & 5.91   & 3.66   & 17.98  & 29.72  & 22.31  & 100.00 \\   
& TIM    & 6.91   & 3.83   & 20.31  & 21.48  & 17.06  & 100.00 \\   
& DIM    & 12.15  & 5.74   & 28.39  & 37.88  & 33.13  & 100.00 \\   
& SIM    & 12.23  & 5.66   & 27.56  & 37.80  & 31.72  & 100.00 \\   
& Admix  & 16.98  & 7.91   & 30.64  & 41.46  & 41.13  & 99.91  \\   
& Mixup  & 19.23  & 8.65   & 29.47  & 42.38  & 42.54  & 96.83  \\   
& Mixcut & 18.98  & 9.65   & 30.55  & 44.46  & 40.96  & 100.00 \\   
& SSA    & 14.15  & 6.91   & 26.64  & 37.13  & 31.80  & 100.00 \\   
& BSR    & 29.05  & 13.32  & 42.21  & 51.54  & 45.21  & 100.00 \\   
& SFCoT  & 30.39  & 13.40  & 40.04  & 46.87  & 41.93  & 99.91  \\   
& \cellcolor[gray]{0.9}Ours   
& \cellcolor[gray]{0.9}\textbf{44.54} {\footnotesize\textcolor{darkergreen}{(14.15↑)}} 
& \cellcolor[gray]{0.9}\textbf{24.31} {\footnotesize\textcolor{darkergreen}{(10.91↑)}} 
& \cellcolor[gray]{0.9}\textbf{56.36} {\footnotesize\textcolor{darkergreen}{(14.15↑)}} 
& \cellcolor[gray]{0.9}\textbf{69.19} {\footnotesize\textcolor{darkergreen}{(17.65↑)}} 
& \cellcolor[gray]{0.9}\textbf{62.94} {\footnotesize\textcolor{darkergreen}{(17.73↑)}} 
& \cellcolor[gray]{0.9}\textbf{100.00} {\footnotesize\textcolor{darkergreen}{(0.00↑)}} \\

\bottomrule
\end{tabular}

\label{longtab:1}
\end{table*}

\begin{table*}[t]
\centering
\footnotesize
\renewcommand{\arraystretch}{0.8}
\caption{Attack Success Rates (\%) on MTARSI Dataset, ↑ indicates performance improvement over the best baseline.}
\begin{tabular}{@{}>{\centering\arraybackslash}p{1.5cm}|>{\centering\arraybackslash}p{1.5cm}|>{\raggedright\arraybackslash}p{2cm}>{\raggedright\arraybackslash}p{2cm}>{\raggedright\arraybackslash}p{2cm}>{\raggedright\arraybackslash}p{2cm}>{\raggedright\arraybackslash}p{2cm}>{\raggedright\arraybackslash}p{2cm}}

\toprule
Model & Attack & Vgg-16 & Vgg-19 & Res-34 & Res-50 & Dense-121 & IncRes-v2 \\
\midrule

\multirow{13}{*}{Vgg-16} 
  & PGD    & 100.0 & 64.7  & 20.3  & 22.5  & 28.3  & 8.4  \\
  & MIM    & 99.9  & 93.6  & 47.8  & 51.3  & 60.4  & 29.7 \\
  & PIM    & 100.0 & 93.4  & 48.5  & 53.0  & 58.8  & 25.2 \\
  & TIM    & 100.0 & 87.2  & 45.4  & 50.7  & 52.8  & 19.5 \\
  & DIM    & 100.0 & 97.3  & 61.6  & 65.8  & 69.2  & 32.8 \\
  & SIM    & 100.0 & 96.6  & 47.5  & 53.5  & 63.2  & 30.4 \\
  & Admix  & 100.0 & 98.6  & 56.8  & 60.4  & 68.8  & 32.1 \\
  & Mixup  & 99.8  & 97.9  & 50.8  & 54.6  & 66.0  & 30.6 \\
  & Mixcut & 97.3 & 98.2  & 53.1  & 57.4  & 66.4  & 30.7 \\
  & SSA    & 100.0 & 95.3  & 51.5  & 55.2  & 62.7  & 28.3 \\
  & BSR    & 100.0 & 99.4  & 65.6  & 70.1  & 78.0  & 33.2 \\
  & SFCoT  & 100.0 & 99.9  & 77.1  & 81.9  & 87.2  & 34.7 \\
  & \cellcolor[gray]{0.9}Ours 
  & \cellcolor[gray]{0.9}\textbf{100.00} {\footnotesize\textcolor{darkergreen}{(0.00↑)}} 
  & \cellcolor[gray]{0.9}\textbf{100.00} {\footnotesize\textcolor{darkergreen}{(0.10↑)}} 
  & \cellcolor[gray]{0.9}\textbf{90.40} {\footnotesize\textcolor{darkergreen}{(13.30↑)}} 
  & \cellcolor[gray]{0.9}\textbf{91.76} {\footnotesize\textcolor{darkergreen}{(9.86↑)}} 
  & \cellcolor[gray]{0.9}\textbf{94.28} {\footnotesize\textcolor{darkergreen}{(7.08↑)}} 
  & \cellcolor[gray]{0.9}\textbf{52.59} {\footnotesize\textcolor{darkergreen}{(17.89↑)}} \\
\midrule

\multirow{13}{*}{Vgg-19} 
  & PGD    & 61.3  & 100.0 & 18.9  & 20.3  & 27.9  & 7.8  \\
  & MIM    & 94.8  & 99.9  & 48.3  & 52.5  & 59.8  & 25.4 \\
  & PIM    & 94.7  & 100.0 & 50.7  & 51.5  & 58.0  & 22.2 \\
  & TIM    & 85.2  & 100.0  & 44.8  & 52.1  & 52.6  & 15.5 \\
  & DIM    & 98.1  & 99.9 & 59.0  & 63.6  & 69.6  & 30.4 \\
  & SIM    & 97.8  & 100.0 & 52.0  & 58.2  & 73.9  & 25.1 \\
  & Admix  & 99.7  & 100.0  & 62.0  & 64.9  & 70.9  & 26.2 \\
  & Mixup  & 98.6  & 99.1  & 55.5  & 60.3  & 67.0  & 24.5 \\
  & Mixcut & 98.9  & 99.4  & 54.7  & 59.2  & 67.3  & 25.1 \\
  & SSA    & 96.7  & 99.9 & 52.6  & 53.8  & 63.4  & 27.7 \\
  & BSR    & 99.8 & 100.0 & 66.6  & 71.1  & 77.6  & 30.5 \\
  & SFCoT  & 100.0 & 100.0 & 76.7  & 81.5  & 88.5  & 31.2 \\
  & \cellcolor[gray]{0.9}Ours 
  & \cellcolor[gray]{0.9}\textbf{100.00} {\footnotesize\textcolor{darkergreen}{(0.00↑)}} 
  & \cellcolor[gray]{0.9}\textbf{100.00} {\footnotesize\textcolor{darkergreen}{(0.00↑)}} 
  & \cellcolor[gray]{0.9}\textbf{92.13} {\footnotesize\textcolor{darkergreen}{(15.43↑)}} 
  & \cellcolor[gray]{0.9}\textbf{92.97} {\footnotesize\textcolor{darkergreen}{(11.47↑)}} 
  & \cellcolor[gray]{0.9}\textbf{96.11} {\footnotesize\textcolor{darkergreen}{(7.61↑)}} 
  & \cellcolor[gray]{0.9}\textbf{47.71} {\footnotesize\textcolor{darkergreen}{(16.51↑)}} \\
\midrule

\multirow{13}{*}{Res-34} 
  & PGD    & 4.1   & 3.8   & 75.8  & 12.3  & 9.3   & 7.4  \\
  & MIM    & 27.0  & 27.9  & 100.0 & 43.4  & 47.9  & 31.2 \\
  & PIM    & 18.6  & 20.2  & 99.9  & 40.9  & 41.1  & 20.1 \\
  & TIM    & 3.1  & 2.4  & 100.0 & 19.6  & 16.3  & 2.7 \\
  & DIM    & 44.9  & 45.5  & 99.9  & 68.4  & 68.3  & 33.8 \\
  & SIM    & 38.5  & 38.9  & 99.8 & 48.1  & 51.3  & 25.8 \\
  & Admix  & 49.4  & 49.1  & 100.0  & 64.3  & 66.2  & 31.3 \\
  & Mixup  & 40.9  & 40.8  & 92.9  & 55.4  & 60.4  & 28.3 \\
  & Mixcut & 38.2  & 40.9  & 99.7  &55.4  & 59.4  & 28.5 \\
  & SSA    & 51.4  & 51.8  & 99.9 & 65.5  & 64.6  & 29.8 \\
  & BSR    & 68.7  & 68.0  & 100.0 & 88.8  & 84.6  & 35.8 \\
  & SFCoT  & 77.2  & 75.5  & 100.0 & 91.6  & 86.0  & 32.4 \\
  & \cellcolor[gray]{0.9}Ours 
  & \cellcolor[gray]{0.9}\textbf{94.38} {\footnotesize\textcolor{darkergreen}{(17.18↑)}} 
  & \cellcolor[gray]{0.9}\textbf{93.65} {\footnotesize\textcolor{darkergreen}{(18.15↑)}} 
  & \cellcolor[gray]{0.9}\textbf{100.00} {\footnotesize\textcolor{darkergreen}{(0.00↑)}} 
  & \cellcolor[gray]{0.9}\textbf{97.74} {\footnotesize\textcolor{darkergreen}{(6.14↑)}} 
  & \cellcolor[gray]{0.9}\textbf{99.10} {\footnotesize\textcolor{darkergreen}{(13.10↑)}} 
  & \cellcolor[gray]{0.9}\textbf{66.91} {\footnotesize\textcolor{darkergreen}{(31.11↑)}} \\
\midrule

\multirow{13}{*}{Res-50} 
  & PGD    & 4.9   & 5.0   & 11.5  & 66.6  & 8.7   & 7.3  \\
  & MIM    & 22.1  & 23.4  & 41.2  & 99.8  & 45.9  & 25.9 \\
  & PIM    & 15.8  & 17.1  & 37.7  & 99.2  & 39.0  & 17.1 \\
  & TIM    & 2.7  & 2.9  & 16.0  & 99.0  & 15.9  & 1.7 \\
  & DIM    & 42.9  & 43.0  & 67.4  & 99.8  & 71.6  & 28.0 \\
  & SIM    & 31.2  & 29.2  & 38.7  & 93.4  & 46.2  & 21.3 \\
  & Admix  & 43.0  & 42.2  & 55.7  & 99.6  & 66.7  & 25.9 \\
  & Mixup  & 34.6  & 34.9  & 46.3  & 94.7  & 57.8  & 23.7 \\
  & Mixcut & 33.7  & 34.3  & 45.1  & 98.6  & 57.4  & 23.6 \\
  & SSA    & 50.1  & 52.2  & 65.6  & 99.7 & 64.6  & 23.7 \\
  & BSR    & 59.8  & 61.0  & 81.2  & 100.0 & 82.2  & 28.3 \\
  & SFCoT  & 76.0  & 75.7  & 91.1  & 100.0 & 89.6  & 28.6 \\
  & \cellcolor[gray]{0.9}Ours 
  & \cellcolor[gray]{0.9}\textbf{95.22} {\footnotesize\textcolor{darkergreen}{(19.22↑)}} 
  & \cellcolor[gray]{0.9}\textbf{93.49} {\footnotesize\textcolor{darkergreen}{(17.79↑)}} 
  & \cellcolor[gray]{0.9}\textbf{99.42} {\footnotesize\textcolor{darkergreen}{(8.32↑)}} 
  & \cellcolor[gray]{0.9}\textbf{100.00} {\footnotesize\textcolor{darkergreen}{(0.00↑)}} 
  & \cellcolor[gray]{0.9}\textbf{99.16} {\footnotesize\textcolor{darkergreen}{(9.56↑)}} 
  & \cellcolor[gray]{0.9}\textbf{60.88} {\footnotesize\textcolor{darkergreen}{(32.28↑)}} \\
\midrule

\multirow{13}{*}{\parbox{1.5cm}{\centering Dense-121}} 
  & PGD    & 5.4   & 5.2   & 12.6   & 12.7  & 78.3  & 8.2  \\
  & MIM    & 26.9  & 28.1  & 38.0  & 38.6  & 99.9  & 28.9 \\
  & PIM    & 21.8  & 22.0  & 37.5  & 40.0  & 99.9  & 19.2 \\
  & TIM    & 3.4  & 3.5  & 14.5 & 18.9  & 99.9  & 3.0 \\
  & DIM    & 48.0  & 49.1  & 61.6  & 63.4  & 99.9  & 33.4 \\
  & SIM    & 32.0  & 33.7  & 37.8  & 41.6  & 99.8 & 25.2 \\
  & Admix  & 43.9  & 47.9  & 54.9  & 57.7  & 100.0  & 30.8 \\
  & Mixup  & 34.8  & 38.5  & 45.1  & 49.4  & 96.0  & 28.3 \\
  & Mixcut & 34.4  & 37.8  & 46.3  & 47.5  & 91.6  & 28.0 \\
  & SSA    & 49.1  & 55.1  & 60.3  & 63.9  & 99.9 & 29.2 \\
  & BSR    & 64.3  & 64.2  & 73.2  & 79.7  & 100.0 & 35.0 \\
  & SFCoT  & 81.7  & 82.3  & 88.4  & 91.9  & 100.0 & 33.0 \\
  & \cellcolor[gray]{0.9}Ours 
  & \cellcolor[gray]{0.9}\textbf{94.96} {\footnotesize\textcolor{darkergreen}{(13.26↑)}} 
  & \cellcolor[gray]{0.9}\textbf{96.64} {\footnotesize\textcolor{darkergreen}{(14.34↑)}} 
  & \cellcolor[gray]{0.9}\textbf{99.00} {\footnotesize\textcolor{darkergreen}{(10.60↑)}} 
  & \cellcolor[gray]{0.9}\textbf{97.11} {\footnotesize\textcolor{darkergreen}{(5.21↑)}} 
  & \cellcolor[gray]{0.9}\textbf{100.00} {\footnotesize\textcolor{darkergreen}{(0.00↑)}} 
  & \cellcolor[gray]{0.9}\textbf{67.22} {\footnotesize\textcolor{darkergreen}{(32.22↑)}} \\
\midrule

\multirow{13}{*}{\parbox{1.5cm}{\centering IncRes-v2}} 
  & PGD    & 2.8   & 2.7   & 5.3   & 6.2   & 4.6   & 58.7 \\
  & MIM    & 4.3   & 5.0   & 7.8   & 8.8   & 7.6   & 100.0 \\
  & PIM    & 4.7   & 4.9   & 8.0   & 9.1   & 8.5   & 100.0 \\
  & TIM    & 1.3   & 1.1   & 3.3  & 4.2  & 3.6  & 100.0 \\
  & DIM    & 6.6   & 7.8   & 12.8   & 15.6   & 13.9   & 100.0 \\
  & SIM    & 4.3   & 5.4   & 6.2   & 8.3   & 7.6   & 62.3 \\
  & Admix  & 4.9   & 5.5   & 8.5   & 9.7   & 8.4   & 87.4 \\
  & Mixup  & 3.7   & 4.3   & 6.9   & 6.7  & 7.2   & 96.5 \\
  & Mixcut & 4.1   & 4.3   & 6.8  & 6.5  & 7.4  & 96.6 \\
  & SSA    & 4.9   & 5.5  & 9.0  & 10.2  & 8.6  & 98.1 \\
  & BSR    & 7.9   & 9.5  & 13.5  & 16.7  & 14.2  & 100.0 \\
  & SFCoT  & 9.7   & 11.2  & 13.6  & 19.1  & 17.8  & 100.0 \\
  & \cellcolor[gray]{0.9}Ours 
  & \cellcolor[gray]{0.9}\textbf{11.64} {\footnotesize\textcolor{darkergreen}{(1.94↑)}} 
  & \cellcolor[gray]{0.9}\textbf{11.74} {\footnotesize\textcolor{darkergreen}{(0.54↑)}} 
  & \cellcolor[gray]{0.9}\textbf{21.23} {\footnotesize\textcolor{darkergreen}{(7.63↑)}} 
  & \cellcolor[gray]{0.9}\textbf{23.59} {\footnotesize\textcolor{darkergreen}{(4.49↑)}} 
  & \cellcolor[gray]{0.9}\textbf{19.97} {\footnotesize\textcolor{darkergreen}{(2.17↑)}} 
  & \cellcolor[gray]{0.9}\textbf{100.00} {\footnotesize\textcolor{darkergreen}{(0.00↑)}} \\

\bottomrule
\end{tabular}

\label{longtab:2}
\end{table*}

\begin{table*}
\centering
\caption{Comparison of the adversarial examples generated by different methods with the original images.}
\setlength{\tabcolsep}{1pt} % Reduce horizontal spacing between columns
\begin{tabular}{ccccccccccc}
\textbf{Clean RSI} & \textbf{PIM} & \textbf{TIM} & \textbf{DIM} & \textbf{Admix} & \textbf{Mixup} & \textbf{Mixcut} & \textbf{SSA} & \textbf{BSR} & \textbf{SFCoT} & \textbf{Ours} \\ 
% Row 1 of images
\includegraphics[width=0.08\textwidth]{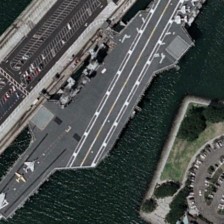} & 
\includegraphics[width=0.08\textwidth]{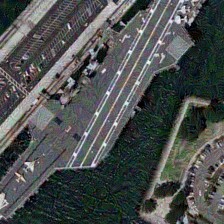} & 
\includegraphics[width=0.08\textwidth]{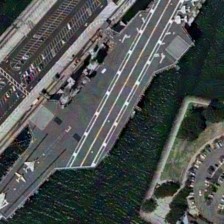} & 
\includegraphics[width=0.08\textwidth]{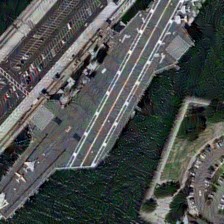} & 
\includegraphics[width=0.08\textwidth]{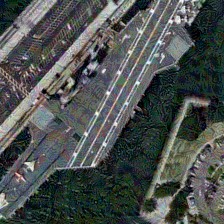} & 
\includegraphics[width=0.08\textwidth]{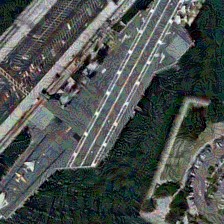} & 
\includegraphics[width=0.08\textwidth]{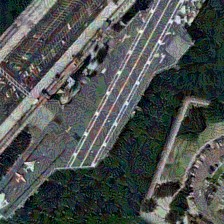} & 
\includegraphics[width=0.08\textwidth]{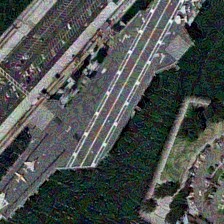} & 
\includegraphics[width=0.08\textwidth]{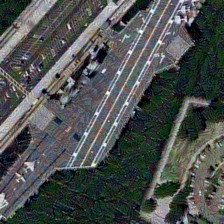} & 
\includegraphics[width=0.08\textwidth]{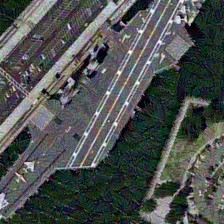} & 
\includegraphics[width=0.08\textwidth]{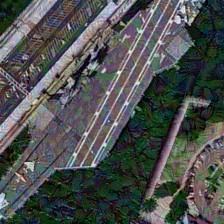} \\ 
% Row 2 of images
\includegraphics[width=0.08\textwidth]{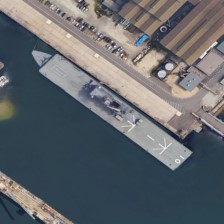} & 
\includegraphics[width=0.08\textwidth]{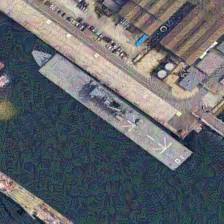} & 
\includegraphics[width=0.08\textwidth]{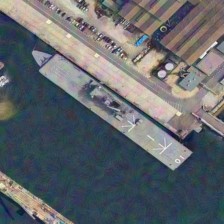} & 
\includegraphics[width=0.08\textwidth]{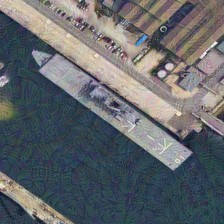} & 
\includegraphics[width=0.08\textwidth]{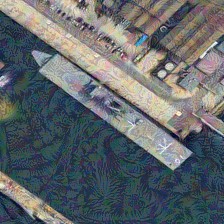} & 
\includegraphics[width=0.08\textwidth]{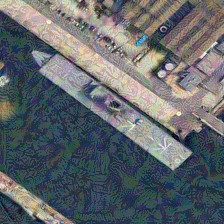} & 
\includegraphics[width=0.08\textwidth]{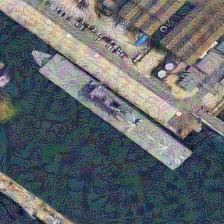} & 
\includegraphics[width=0.08\textwidth]{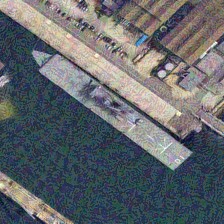} & 
\includegraphics[width=0.08\textwidth]{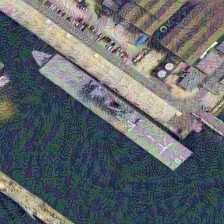} & 
\includegraphics[width=0.08\textwidth]{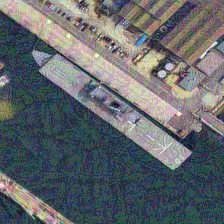} & 
\includegraphics[width=0.08\textwidth]{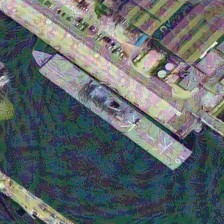} \\ 
% Row 3 of images
\includegraphics[width=0.08\textwidth]{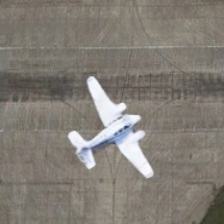} & 
\includegraphics[width=0.08\textwidth]{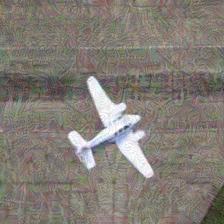} & 
\includegraphics[width=0.08\textwidth]{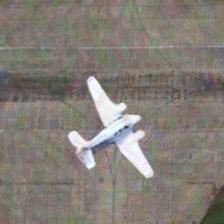} & 
\includegraphics[width=0.08\textwidth]{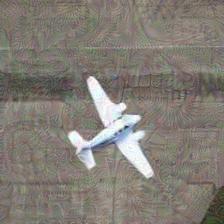} & 
\includegraphics[width=0.08\textwidth]{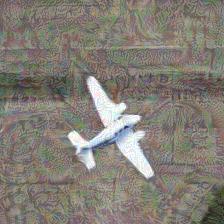} & 
\includegraphics[width=0.08\textwidth]{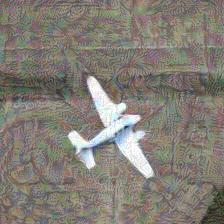} & 
\includegraphics[width=0.08\textwidth]{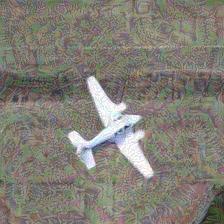} & 
\includegraphics[width=0.08\textwidth]{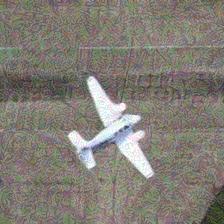} & 
\includegraphics[width=0.08\textwidth]{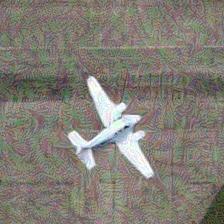} & 
\includegraphics[width=0.08\textwidth]{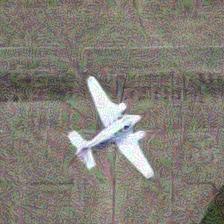} & 
\includegraphics[width=0.08\textwidth]{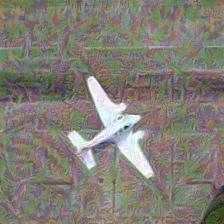} \\ 
% Row 4 of images
\includegraphics[width=0.08\textwidth]{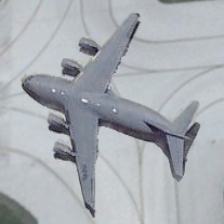} & 
\includegraphics[width=0.08\textwidth]{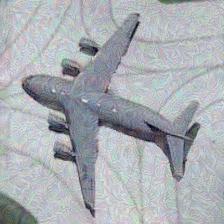} & 
\includegraphics[width=0.08\textwidth]{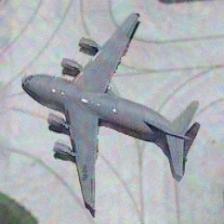} & 
\includegraphics[width=0.08\textwidth]{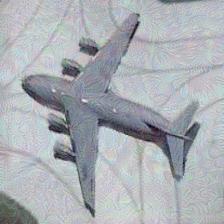} & 
\includegraphics[width=0.08\textwidth]{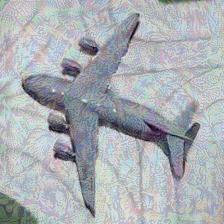} & 
\includegraphics[width=0.08\textwidth]{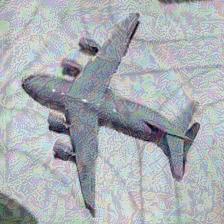} & 
\includegraphics[width=0.08\textwidth]{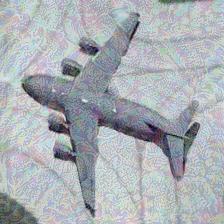} & 
\includegraphics[width=0.08\textwidth]{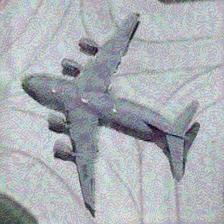} & 
\includegraphics[width=0.08\textwidth]{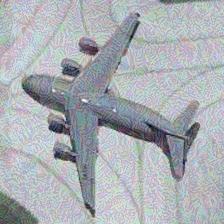} & 
\includegraphics[width=0.08\textwidth]{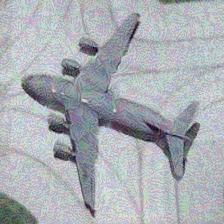} & 
\includegraphics[width=0.08\textwidth]{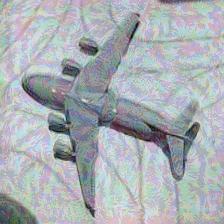} \\ 
\end{tabular}

\label{tab:image_comparison}

\end{table*}
\subsubsection*{2) Computational Complexity}
This paper compares the time complexity of the proposed method with other input transformation-based attack methods. The experiment was conducted using Resnet-34 as the substitute model on the MTARSI dataset. We compared the time consumption and overall attack success rate (OASR) of the methods, where OASR is defined as the average ASR of adversarial examples across multiple black-box models. According to the results shown in Table~\ref{tab:metrics_comparison}, our method reaches an OASR of 90.4, delivering a 38.4\% improvement over Admix while consuming a similar amount of time. These  experimental results show that the proposed method achieves a favorable balance between efficiency and effectiveness.

%We set the parameter $M = 5$, which was kept consistent with other methods. In the quantitative analysis, we set the parameter $M = 5$ to control the number of gradient averaging steps. Since gradient averaging is an effective strategy to mitigate overfitting but incurs additional time cost, we also conducted a comparison of the algorithm’s time complexity with $N = 1$ . 

\begin{table}[ht!]
\centering
\caption{Comparison of Computational Complexity and Overall Attack Success Rate.}
\small % Reduce font size to fit in single column
\setlength{\tabcolsep}{3pt} % Adjust column spacing to fit
\begin{tabular}{lccccccc}
\toprule
 & \text{SIM} & \text{Admix} & \text{SSA} & \text{BSR} & \text{SFCoT}  & \textbf{Ours} \\
\midrule
\textbf{Time (s)} & 785 & 2075 & 453 & 214 & 399  & 2130 \\
\textbf{OSAR} & 40.5 & 52.0 & 52.6 & 69.1 & 72.5 & 90.4 \\
\bottomrule
\end{tabular}
\label{tab:metrics_comparison}
\end{table}
\subsection{Ablation Studies}
To explore the impact of different mixing strategies on attack transferability, we designed an ablation study specifically targeting input transformation components. We propose Local Mix, a local mixing strategy that enhances adversarial examples by mixing different images at localized regions. To verify the contribution of our input transformation to transferability, we further designed a comparison baseline by replacing our transformation pipeline with Admix, where two different images are globally mixed, while keeping all other loss terms unchanged. No Mix refers to a baseline setting where no input transformation or mixing techniques are applied. As shown in the Table~\ref{tab:batch} and Table~\ref{tab:batch2}, our Local Mix strategy achieves higher attack success rates than both Admix and No Mix baselines in most cases. This focused analysis demonstrates the effectiveness of our proposed local mixing approach for improving adversarial transferability.

\begin{table}[ht!]
\centering
\caption{Ablation study of different mix strategies across various models on FGSCR-42.}
\adjustbox{width=\textwidth/2,center}{%
\begin{tabular}{l@{\hskip 6pt}l@{\hskip 8pt}cccc}
\toprule
Model & Mix Strategy  & Vgg-19 & Res-34 & Dense-121 & IncRes-v2 \\
\midrule
\multirow{3}{*}{Vgg-19} 
  & Local-Mix  & 100.00 & 97.58 & 98.33 & 53.03 \\
  & Admix   & 100.00 & 96.39  & 97.45 &52.83  \\
  & No-Mix   & 100.00 & 95.50  & 97.00 & 51.20  \\ 
\midrule
\multirow{3}{*}{Res-34} 
  & Local-Mix  & 95.50 & 100.00  & 99.58 & 68.44 \\
  & Admix  & 92.45 & 100.00  & 99.17 & 67.26 \\
  & No-Mix   & 88.92 & 100.00  & 98.83 & 63.28  \\ 
\midrule
\multirow{3}{*}{Dense-121} 
  & Local-Mix  & 96.75 & 99.66 & 100.00 & 69.60  \\
  & Admix  & 93.76 & 97.94  & 99.83 & 68.00 \\
  & No-Mix   & 91.67 & 97.33  & 100.00 & 63.86  \\ 
\bottomrule
\end{tabular}
}
\label{tab:batch}
\end{table}

\begin{table}[ht!]
\centering
\caption{Ablation study of different mix strategies across various models on MTARSI.}
\adjustbox{width=\textwidth/2,center}{%
\begin{tabular}{l@{\hskip 6pt}l@{\hskip 8pt}cccc}
\toprule
Model & Mix Strategy  & Vgg-19 & Res-34 & Dense-121 & IncRes-v2 \\
\midrule
\multirow{3}{*}{Vgg-19} 
  & Local-Mix  & 100.00 & 92.13  & 96.11 & 47.71 \\
  & Admix   & 100.00 & 90.50  & 96.80 & 45.51  \\
  & No-Mix   & 100.00 & 88.98  & 95.85 & 41.95  \\  
\midrule
\multirow{3}{*}{Res-34} 
  & Local-Mix  & 93.65 & 100.00  & 99.91 & 66.91 \\
  & Admix  & 90.03 & 100.00  & 96.38 & 57.89 \\
  & No-Mix   & 87.46 & 100.00  & 97.01 & 55.79  \\ 
\midrule
\multirow{3}{*}{Dense-121} 
  & Local-Mix  & 96.64 & 99.00 & 100.00 & 67.22  \\
  & Admix  & 96.59 & 98.37  & 100.00 & 62.71 \\
  & No-Mix   & 94.17 & 95.64  & 100.00 & 58.04  \\ 
\bottomrule
\end{tabular}
}
\label{tab:batch2}
\end{table}

Furthermore, to analyze the contribution of the proposed loss function to attack performance, we conducted an additional ablation study on the loss components. Our method jointly optimizes the gradient direction using both a logit loss and a perturbation smoothing loss. Table~\ref{tab:ablation_study} and Table~\ref{tab:ablation_study2} illustrates the attack success rates when using Dense-121 and Inception-ResNet-v2 as substitute models against various target models. Specifically, "All" denotes the full loss combination, "w/o Smo." indicates the removal of the perturbation smoothing loss and "CE only" refers to using the cross-entropy loss. The results show that removing the perturbation smoothing loss leads to a significant performance drop, especially when attacking Inception-ResNet-v2, highlighting its importance in enhancing transferability to heterogeneous models. When using only cross-entropy loss, performance further degrades, demonstrating its limited effectiveness in improving transferability. These results demonstrate the superiority of the proposed loss function over traditional cross entropy loss. The logit loss helps avoid gradient vanishing and guides more effective perturbation directions, while the perturbation smoothing loss suppresses high-frequency noise, thereby enhancing adversarial transferability across diverse model architectures.
\begin{table}[ht!]
\centering
\caption{Ablation study of different loss function components when using Inception-ResNet-v2 as  substitute model on FGSCR-42 dataset.}
\small % Reduce font size to fit in single column
 % Adjust column spacing to fit
 \adjustbox{width=\textwidth/2,center}{%
\begin{tabular}{ccccccc}
\toprule
\text{Loss} & \text{Vgg-16} & \text{Vgg-19} & \text{Res-34} & \text{Res-50} & \text{Dense-121} \\
\midrule
\text{All}           & 44.54 & 24.31 & 56.36 & 69.19 & 62.94 \\
\text{w/o Smo.}   & 34.47 & 18.56 & 48.95 & 62.36 & 56.53 \\
\text{only CE}       & 31.14 & 16.56 & 48.79 & 62.19 & 57.78 \\
\bottomrule
\end{tabular}
}
\label{tab:ablation_study}
\end{table}

\begin{table}[ht!]
\centering
\caption{Ablation study of different loss function components when using Dense-121 as substitute model on FGSCR-42 dataset.}
\small % Reduce font size to fit in single column
 % Adjust column spacing to fit
\adjustbox{width=\textwidth/2,center}{%
\begin{tabular}{ccccccc}
\toprule
\text{Loss} & \text{Vgg-16} & \text{Vgg-19} & \text{Res-34} & \text{Res-50} & \text{IncRes-v2} \\
\midrule
\text{All}           & 97.08 & 96.83 & 99.50 & 99.66 & 69.85 \\
\text{w/o Smo.}   & 96.16 & 96.25 & 99.41 & 99.83 & 60.53 \\
\text{only CE}       & 91.00 & 88.17 & 98.66 & 99.58 & 59.45 \\
\bottomrule
\end{tabular}
}
\label{tab:ablation_study2}
\end{table}

\subsection{Parameter Analysis}
This section analyzes the parameters used in the experiments, which include the transformation times \( M \), the loss weight \( \lambda \), and the image mixing ratio \( \eta \). The parameter experiments were conducted on FGSCR-42 using ResNet-34 as the substitute model.

%Fig.~\ref{fig:N} illustrates the impact of different \( N \) values on the attack success rate, where \( N \) controls the degree of gradient smoothness. Observing the chart, it can be seen that when \( N \) is less than or equal to 5, the ASR gradually increases as \( N \) rises. However, when \( N \) exceeds 5, although the attack performance continues to improve in the Inception-ResNet-v2 model, the performance improvement in other target models becomes less significant. This indicates that the contribution of gradient averaging to the attack success rate is limited, and excessive smoothing consumes more time while yielding diminishing returns. Balancing time cost and performance, the experiment selected \( N = 5 \) as the optimal setting.

\begin{figure}[t]
  \centering
    \includegraphics[height=6.5cm]{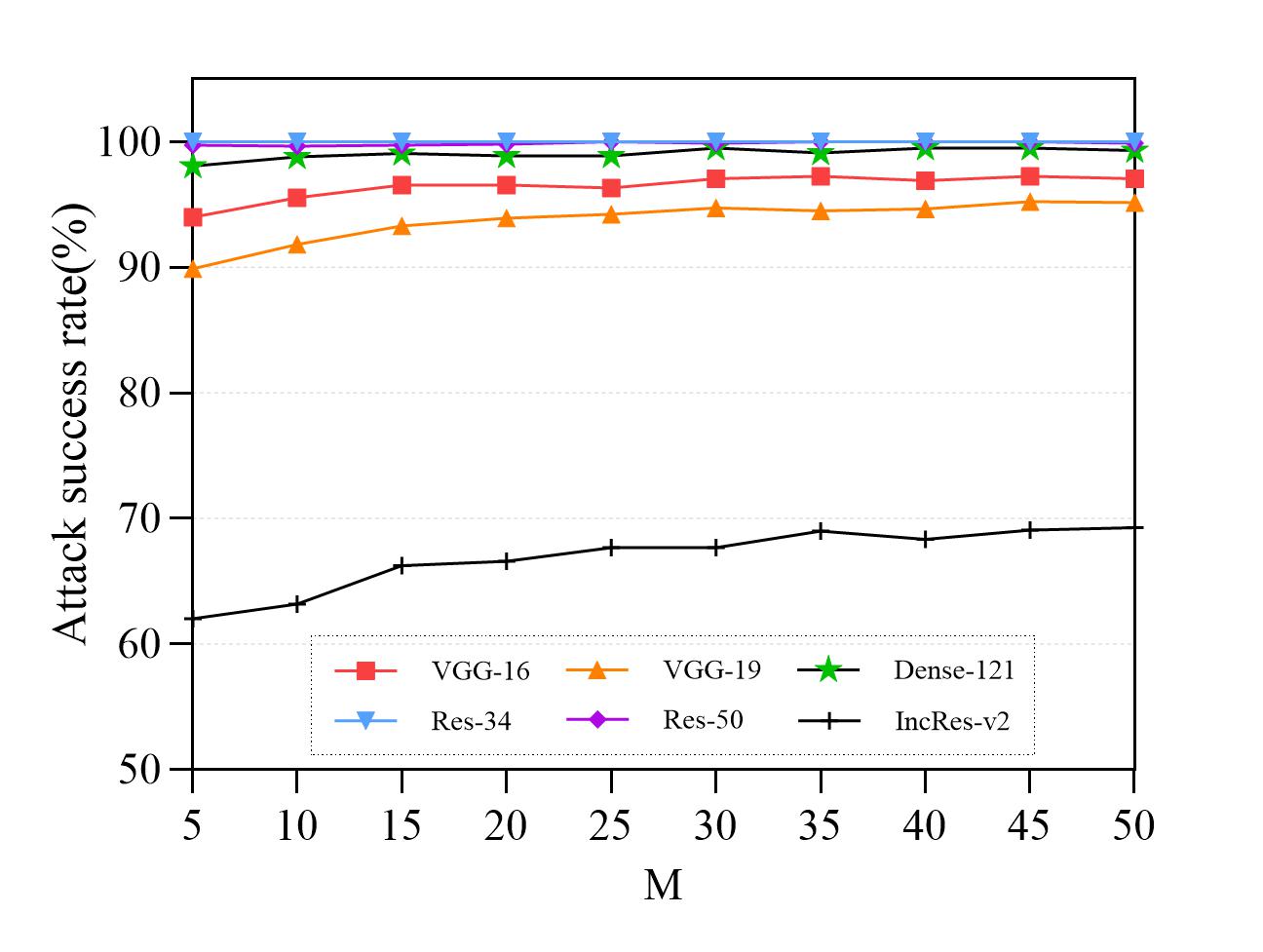} % 替换为你的图片文件名
  \caption{Attack success rate with different settings M .}
  \label{fig:M}
\end{figure}

Fig.~\ref{fig:M}  illustrates the effect of the input transformation times \( M \) on the attack success rate across different models. As \( M \) increases from 5 to 25, the attack success rate consistently improves, especially for models like Inception-ResNet-V2  that are harder to attack in the black-box setting. Beyond \( M \)=25, the improvements tend to plateau for most models, suggesting diminishing returns. Considering this trade-off between attack effectiveness and computational overhead, we choose \( M \)=25 as a balanced setting. It ensures strong transferability while avoiding unnecessary computation from excessive transformations.

%Fig.~\ref{fig:NM} demonstrates the effect of \( M \) values on the attack success rate. From the chart, it is evident that when \( M \) is less than or equal to 5, the attack success rate gradually increases with the rise of \( M \). In particular, the attack success rate performs most ideally when \( M \) is set to 5. As \( M \) continues to increase beyond this point, the attack success rate stabilizes and does not show significant further improvement. This suggests that increasing \( M \) helps construct more transformed image sets, thereby enhancing the transferability of the attack, but excessively high \( M \) values may lead to diminishing returns. Considering overall efficiency, we set \( M = 5 \).

 Fig.~\ref{fig:yita} demonstrates the effect of varying the mixing ratio \( \eta \) on attack success rate. When \( \eta = 0 \), the method reverts to not applying any mixing, resulting in the lowest attack success rate. The highest attack success rate is achieved when \( \eta = 0.5 \), indicating that a moderate mixing ratio optimizes performance.

\begin{figure}[t]
  \centering
    \includegraphics[height=6.5cm]{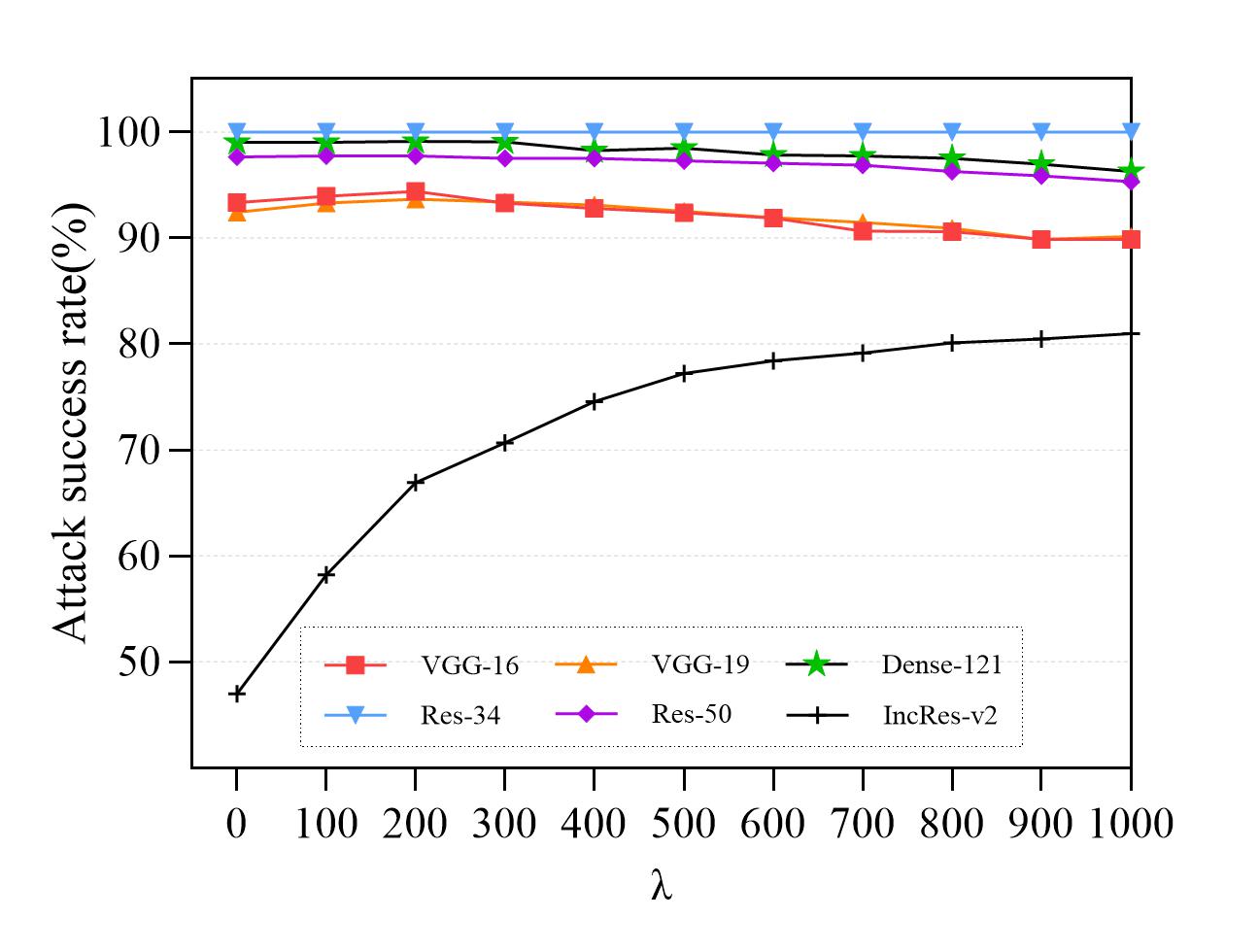} % 替换为你的图片文件名
    \label{fig:ratio}
  \caption{Attack success rate with different settings of $\lambda$ .}
  \label{fig:lambda}
\end{figure}

\begin{figure}[t]
  \centering
    \includegraphics[height=6.5cm]{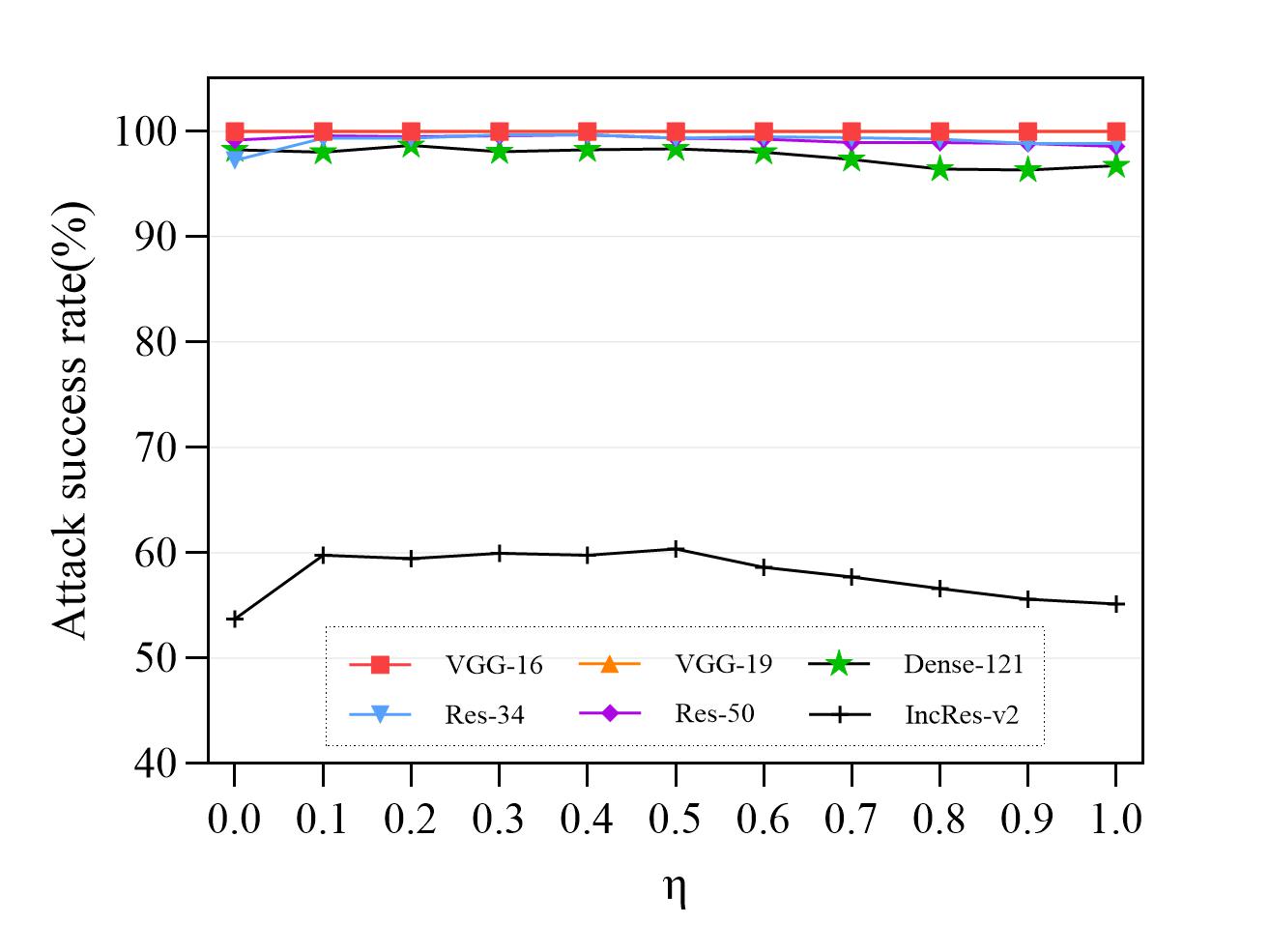} % 替换为你的图片文件名
    \label{fig:ratio}
  \caption{Attack success rate with different settings of $\eta$ .}
  \label{fig:yita}
\end{figure}

To determine the optimal $\lambda$ parameter, we conducted sensitivity experiments on the MTARSI dataset using ResNet-34 as the white-box surrogate model, evaluating the impact of different $\lambda$ values on attack success rates (ASR) across six target models. As shown in the Fig.~\ref{fig:lambda}, VGG series, ResNet series, and DenseNet-121 models maintained high ASR above 90\% from $\lambda=0$ to $\lambda=200$, but performance began to decline when $\lambda>200$. In contrast, Inception-ResNet-V2  exhibited the opposite trend: ASR was only 47\% at $\lambda=0$ but improved significantly as $\lambda$ increased, reaching 75\% at $\lambda=400$. This phenomenon may be attributed to Inception-ResNet-V2's unique network architecture. Considering the overall performance across all models, we selected $\lambda=200$ as the optimal parameter, which maintains high attack success rates for most models while avoiding performance degradation at higher $\lambda$ values.

\section{Conclusion}
This paper proposes a novel adversarial attack method to address the challenge of generating transferable adversarial examples in remote sensing object recognition. Unlike existing mixing based attack methods to enhance transferability, which either perform global blending or directly exchange a region in the images, we present a local mixing strategy to generate diverse yet semantically consistent inputs during the generation of adversarial examples. The proposed local mixing method merely blends local regions of two images to preserve as much of the global semantic information as possible. Moreover, unlike traditional cross-entropy loss, we design an untargeted attack loss based on logits optimization, which mitigates the gradient vanishing problem inherent to cross entropy loss and constrains the high-frequency components generated during the gradient update process, further enhancing transferability across heterogeneous architectures. Experimental results on the FGSCR-42 and MTARSI datasets show that our method outperforms existing techniques in both attack performance and transferability, effectively improving the attack's effectiveness across multiple model architectures.

\bibliographystyle{unsrt}
\bibliography{reference}

\end{document}